\documentclass[pdflatex,sn-basic,Numbered]{sn-jnl}

\usepackage[utf8]{inputenc}
\usepackage{listings}
\usepackage{graphicx}
\usepackage{mathtools}
\usepackage{amsmath}
\usepackage{amssymb}
\usepackage[section]{placeins}
\usepackage[caption=false]{subfig}
\usepackage{hyperref}
\usepackage{bookmark}
\usepackage{url}
\usepackage{float}
\usepackage{changepage}
\usepackage{stackengine,scalerel}

\newcommand\overstar[1]{\ThisStyle{\ensurestackMath{%
\setbox0=\hbox{$\SavedStyle#1$}%
\stackengine{0pt}{\copy0}{\kern.2\ht0\smash{\SavedStyle*}}{O}{c}{F}{T}{S}}}}

\newcommand{\symbolnomenclature}[2]{\indent\parbox{30mm}{#1}{#2\par}}

\begin{document}

\title[Integration of deep generative Anomaly Detection algorithm in high-speed industrial line]{Integration of deep generative Anomaly Detection algorithm in high-speed industrial line}

\author*[1,2]{\fnm{Niccolò} \sur{Ferrari}}\email{niccolo.ferrari@unife.it}
\author[2]{\fnm{Nicola} \sur{Zanarini}}
\author[1]{\fnm{Michele} \sur{Fraccaroli}}
\author[1]{\fnm{Alice} \sur{Bizzarri}}
\author[1]{\fnm{Evelina} \sur{Lamma}}

\affil*[1]{\orgdiv{Department of Engineering}, \orgname{University of Ferrara}, \orgaddress{\street{Via Saragat 1}, \city{Ferrara}, \postcode{44122}, \country{Italy}}}
\affil[2]{\orgname{Bonfiglioli Engineering}, \orgaddress{\street{Via Amerigo Vespucci 20}, \city{Ferrara}, \postcode{44124}, \country{Italy}}}

\abstract{Industrial visual inspection in pharmaceutical production requires high accuracy under strict constraints on cycle time, hardware footprint, and operational cost. Manual inline inspection is still common, but it is affected by operator variability and limited throughput. Classical rule-based computer vision pipelines are often rigid and difficult to scale to highly variable production scenarios. To address these limitations, we present a semi-supervised anomaly detection framework based on a generative adversarial architecture with a residual autoencoder and a dense bottleneck, specifically designed for online deployment on a high-speed Blow-Fill-Seal (BFS) line. The model is trained only on nominal samples and detects anomalies through reconstruction residuals, providing both classification and spatial localization via heatmaps. The training set contains 2,815,200 grayscale patches. Experiments on a real industrial test kit show high detection performance while satisfying timing constraints compatible with a 500 ms acquisition slot.}


\maketitle

\section{Acknowledgments}
	\label{sec: acknowledgments}
	{
		The authors would like to thank Bonfiglioli Engineering for providing a real-case dataset to test the software developed in this work.
		The first author is supported by an industrial PhD funded by Bonfiglioli Engineering, Ferrara, Italy.
		Alice Bizzarri is supported by a National PhD funded by Politecnico di Torino, Torino, Italy and Università di Ferrara, Ferrara, Italy.
	}

{
	\section{Nomenclature}
		\label{sec: nomenclature}
		{
			\noindent
			
			\symbolnomenclature{AE}{AutoEncoder}
			\symbolnomenclature{VAE}{Variational AutoEncoder}
			\symbolnomenclature{CNN}{Convolutional Neural Network}
			\symbolnomenclature{RNN}{Recurrent Neural Network}
			\symbolnomenclature{LSTM}{Long Short Time Memory}
			\symbolnomenclature{GAN}{Generative Adversarial Network}
			\symbolnomenclature{Generator}{Generative subnet of the GAN}
			\symbolnomenclature{Discriminator}{Adversarial subnet of the GAN}
			\symbolnomenclature{Discriminative net}{U-Net subsequent to the GAN used for segmentation}
			\symbolnomenclature{CRAE}{fully-Convolutional Residual AutoEncoder}
			\symbolnomenclature{DRAE}{Dense-bottleneck Residual AutoEncoder}
			\symbolnomenclature{AUROC}{Area Under the Receiver Operating Characteristic}
			\symbolnomenclature{ROI}{Region Of Interest}
			\symbolnomenclature{SSIM}{Structural Similarity Index Measure}
			\symbolnomenclature{BFS}{Blow Fill Seal}
		}
}

	\section{Introduction}
		\label{sec: intro}
		{
			Industrial applications increasingly require machine learning solutions for demanding tasks such as inline visual anomaly detection. In real deployments, models must satisfy stringent constraints on hardware resources and throughput while preserving high inspection accuracy. As a consequence, technical feasibility is as relevant as raw model performance.
			
			Anomaly detection for machine vision is particularly relevant in the pharmaceutical sector, where non-destructive quality controls must be both accurate and fast. For this reason, pharmaceutical manufacturers are investing in integrating deep learning systems into production-line automation. In practice, system design must balance the following constraints:
			\begin{enumerate}
				\item Accuracy (missed rejects and false rejects)
				\item Hardware constraints
				\item Total cost of ownership
			\end{enumerate}
			Accuracy has a dual impact: it directly affects patient safety and, in parallel, business outcomes. Hardware and cost are correlated but not equivalent; hardware choices influence footprint, maintainability, and integration complexity in addition to direct expenditure.
			
			In many production sites, quality control still relies on manual inspection. This process is vulnerable to non-systematic errors, including oversight and attention loss, and therefore limits both consistency and throughput. Cosmetic inspection and particle detection are especially challenging because decisions rely on perceptual cues that are difficult to formalize analytically.
			
			Classical procedural algorithms are typically developed ad hoc from limited sets of compliant and anomalous examples. They rely on handcrafted thresholds, heuristic pattern matching, and many tunable parameters. Although this may appear flexible, the resulting systems are usually tightly coupled to a specific product and image configuration, hence difficult to scale or transfer. In addition, oversimplified analytical models often fail to separate process noise from true defects (e.g., moving bubbles vs. foreign particles), making robust deployment difficult.
			
			Deep learning is currently the state of the art for real-time anomaly detection in machine vision. Supervised classification is mature and widely studied; however, it is most effective when classes are balanced \cite{MOOIJMAN2023109853}.
			
			In industrial production, compliant samples largely outnumber anomalous ones. This imbalance makes supervised training problematic. A practical alternative is semi-supervised learning, where the model is trained only on nominal data and anomalies are identified as out-of-distribution patterns \cite{xu2023deep,anomalib}. Most methods fall into two broad families: \textit{reconstruction-based} and \textit{embedding similarity-based}.
		
		\subsection{Reconstruction-based}
		\label{sec: rec-based}
			This family is based on the assumption that a model trained only on conforming samples cannot faithfully reconstruct anomalous regions. It has been extensively studied because it enables robust reconstruction subspaces without defective training samples. Representative models include autoencoders (AE) \cite{DBLP:journals/corr/abs-2003-05991,DBLP:journals/corr/abs-2201-03898}, variational autoencoders (VAE) \cite{DBLP:journals/corr/abs-1906-02691,https://doi.org/10.48550/arxiv.1312.6114}, generative adversarial networks (GAN) \cite{goodfellow2020generative}, and GAN-based variants such as GANomaly \cite{akcay2018ganomaly,akccay2019skip} and DR{\AE}M \cite{zavrtanik2021draem}. Since anomalous regions are out of distribution with respect to training data, they are typically reconstructed poorly and can be detected by thresholding residuals or similarity measures such as SSIM \cite{wang2004image}.
			These architectures are usually computationally heavy in both training and inference, but they produce compact and manageable latent representations.
		
		\subsection{Embedding similarity-based}
		\label{sec: emb-sim-based}
			This family uses deep neural networks, often pre-trained on large-scale datasets, to extract embedding vectors that summarize visual content.
			The underlying idea is that expressive backbones, such as residual networks \cite{DBLP:journals/corr/HeZRS15}, can provide robust feature representations even when the target dataset is limited. Regularity can then be modeled with methods such as PaDiM \cite{defard2021padim}, PatchCore \cite{roth2022towards,muhr2023probabilistic}, and normalizing-flow approaches including FastFlow \cite{yu2021fastflow}, PNI \cite{bae2023pni}, and MSFlow \cite{zhou2023msflow}. In many formulations, nominal data are approximated with multivariate Gaussian models \cite{articleStatistical}. The main drawback is interpretability: the anomaly score is a distance in feature space rather than a direct reconstruction error \cite{defard2021padim}. In addition, memory requirements can become prohibitive on embedded or on-board hardware because latent-space storage often grows with dataset size.
		
		\subsection{Adopted solution}
		\label{sec: sol-work}
			Based on the considerations above, we developed an inline real-time anomaly detection system for cosmetic inspection of BFS plastic vial strips filled with liquid. The main objective was not only high detection accuracy, but also feasibility under strict project constraints.
		
		The hardware infrastructure is very different between training server and inference computer. For the former:
		\begin{itemize}
			\item Intel\textsuperscript{\textregistered} Xeon\textsuperscript{\textregistered} Silver 4216 CPU @ 2.10 GHz as CPU
			\item 64GB of DDR4 Synchronous @ 3200 MHz as system memory
			\item Nvidia\textsuperscript{\textregistered} A100 with 40GB of VRAM as GPU.
		\end{itemize}
		For the latter:
		\begin{itemize}
			\item Intel\textsuperscript{\textregistered} Xeon\textsuperscript{\textregistered} E-2278GE CPU @ 3.30 GHz as CPU
			\item 32GB of DDR4 Synchronous @ 3200 MHz as system memory
			\item Nvidia\textsuperscript{\textregistered} A4500 with 20GB of VRAM as GPU.
		\end{itemize}
		 
			The proposed architecture is adapted from previous work \cite{GRDNetArticle} to satisfy hardware limits and acceptance-test constraints. The model uses a single GAN with a residual autoencoder (RAE) and a dense bottleneck (DRAE). As in DRAEM, Perlin noise is randomly superimposed on input images during training, not to synthesize defects, but to improve denoising behavior and reduce the tendency of the model to copy the input.
			
			The training dataset contains 2,815,200 images obtained from 782 strips of 5 vials each. Every strip was acquired 10 times with 16 frames per acquisition, plus 2 ranked images. Each image was divided into 20 patches: 5 vial regions, each split into 4 sub-regions.
		
		{
			\medskip
				In summary, the contribution of this work is:
				\begin{enumerate}
					\item A generative network based on a residual autoencoder (RAE) \cite{9373350,Zini_2020,Le2023} with a fully-connected bottleneck, embedded in a GAN architecture \cite{akcay2018ganomaly}.
					\item A preprocessing pipeline that partitions each vial into four logical regions for patch-level analysis.
					\item Training-time augmentation on nominal samples, including randomized Perlin-noise masking to improve robustness against out-of-distribution perturbations.
					\item A multi-stage evaluation protocol from raw patches to vial- and run-level aggregation, aligned with industrial acceptance criteria.
					\item Online deployment of the inference pipeline in the machine control software through C++ TensorFlow APIs.
				\end{enumerate}
			}
		}

	\section{Related Work}
		\label{sec: related}
		{
			The implemented approach is primarily based on GRD-Net \cite{GRDNetArticle}, which is inspired by DR\AE M \cite{zavrtanik2021draem}. This family belongs to reconstruction-based anomaly detection, where models reconstruct nominal inputs and expose out-of-distribution regions through reconstruction mismatch \cite{dimattia2021survey, XIA2022497}. Representative architectures include autoencoders (AE) \cite{DBLP:journals/corr/abs-2003-05991, DBLP:journals/corr/abs-2201-03898, bergmann2018improving, gong2019memorizing}, variational autoencoders (VAE) \cite{DBLP:journals/corr/abs-1906-02691, venkataramanan2020attention}, and GAN-based methods \cite{akcay2018ganomaly, akccay2019skip, pidhorskyi2018generative, sabokrou2018adversarially}.
			In these methods, the encoder maps the image into a compact latent space \cite{DBLP:journals/corr/abs-2003-05991}. SSIM \cite{bergmann2018improving} and pixel-wise reconstruction errors \cite{bergmann2019mvtec} are commonly used both for training and for anomaly scoring at inference. Transformer-based reconstruction models have also been proposed \cite{https://doi.org/10.1002/int.22974, Mishra_2021}, including hybrid convolutional-transformer designs \cite{esser2021taming}. A key advantage of reconstruction-based methods is interpretability through residual maps; a typical drawback is computational cost. Contrastive extensions have shown additional performance gains \cite{tian2021divide}.
			
			A second major family is \emph{embedding similarity-based} anomaly detection. These approaches extract feature vectors for image-level anomaly detection \cite{rippel2021modeling, bergman2020deep} or patch-level localization \cite{napoletano2018anomaly}. Compared with reconstruction-based methods, they are often lighter at inference because they mainly require an encoder. However, they are less interpretable: scores are distances between inferred embeddings and the nominal embedding distribution. Early work such as SPADE \cite{cohen2020sub} modeled normality with hypersphere-based representations. Later methods include PaDiM \cite{defard2021padim}, PatchCore \cite{roth2022towards}, and normalizing-flow variants \cite{dinh2016density, bae2023pni, yu2021fastflow}. These techniques can deliver high throughput, but memory-bank requirements may become prohibitive on constrained industrial hardware. Moreover, reconstruction-based methods can better capture certain logical defects, as discussed by Batzner et al. \cite{batzner2023efficientad}.
		}

	\section{Methods}
		\label{sec: methods}
		{
			To describe our approach, we first summarize the GRD-Net \cite{GRDNetArticle} architecture.
		
		\subsection{GRD-Net}
		\label{sec: grd-net}
		{
				Inspired by DR\AE M \cite{zavrtanik2021draem}, GRD-Net \cite{GRDNetArticle} is composed of two networks. The first is a GAN with a fully-convolutional residual autoencoder (CRAE). The second is a U-Net with skip connections that takes the original image $X$ and the reconstructed image $\hat{X}$ as input.
				During training, a perturbation function $P_q$ superimposes Perlin noise on $X$ with probability $q$, generating a perturbed image $X_n$ and a binary map $M$. In $M$, perturbed pixels are set to $1.0$ and unperturbed pixels to $0.0$. The original $X$ is unchanged, while $X_n$ is generated as:
			\begin{equation}
				X_n, M = P_q(X).
			\end{equation}
			
			\begin{figure}
				\centering
				\subfloat[Original vial region image ($X$)]{\includegraphics[scale=0.4]{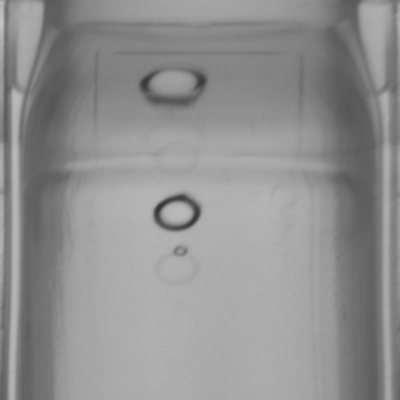}}
				\hspace{0.25cm}\subfloat[Original vial region image with Perlin noise $X_{n}$]{\includegraphics[scale=0.4]{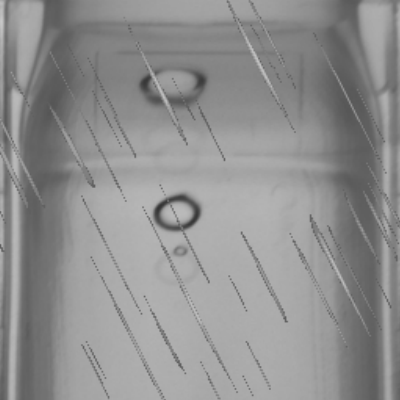}}
				\hspace{0.25cm}\subfloat[The generated Perlin noise standalone $N$]{\includegraphics[scale=0.4]{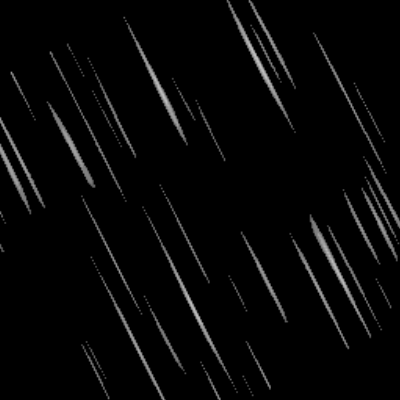}}
				\hspace{0.25cm}\subfloat[The map of the noise $M$]{\includegraphics[scale=0.4]{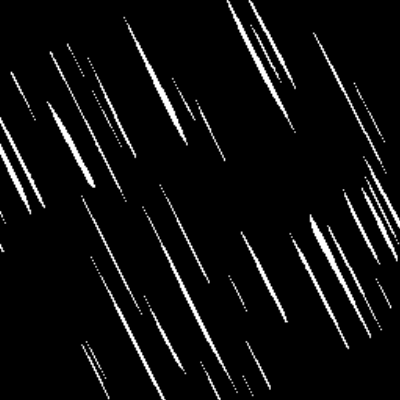}}
					\caption{(a) Original vial-region image $X$. (b) Same image with superimposed Perlin noise $X_n$. (c) Isolated Perlin noise map $N$. (d) Binary mask $M$ of the superimposed region.}
				\label{fig:pills_ex}
			\end{figure}
			
				In this design, perturbation acts as an augmentation mechanism and forces the autoencoder to solve a denoising task in addition to feature compression. It also produces the supervision map used by the second network for anomaly-region prediction.
				
				The superimposed noise forces the network to reconstruct masked content and therefore emphasizes essential structure learning. Similar principles were investigated, for example, in masked autoencoding approaches \cite{DBLP:journals/corr/abs-2111-06377}.
				
				In the initial project, a Region of Interest (ROI) attention map was also used during training to focus learning on predefined areas that vary across samples.
			
			\subsubsection{Residual Autoencoders}
			\label{sec: autoencoders}
			{
					An autoencoder (AE) is trained to reconstruct an input image, ideally filtering out nuisance noise. It is composed of an encoder (E), which maps input features into a latent representation $z$, and a decoder (D), which reconstructs the input from $z$.
					To succeed, the model must preserve informative structure while discarding non-relevant variability. This is typically encouraged by constraining the latent dimensionality.
					Residual architectures are widely adopted because they mitigate optimization degradation (e.g., vanishing gradients) \cite{he2015deep}. Although historically dominant in supervised tasks, they are now broadly used in unsupervised and semi-supervised settings \cite{GRDNetArticle,9373350,Zini_2020,Le2023}.
				}
			\subsubsection{Generative Adversarial Networks}
			\label{sec: gans}
			{
					GANs were originally proposed for realistic synthetic data generation \cite{goodfellow2020generative}. They consist of a generator and a discriminator trained in competition: the discriminator distinguishes real from generated samples, while the generator learns to fool it. As in GANomaly \cite{akcay2018ganomaly}, the discriminator in our setting compares reconstructed and original images.
				}
			\subsubsection{GRD-Net architecture}
			\label{sec: grd-net-arch}
			{
					The first component of GRD-Net is the Generator (G), composed of an Encoder (G\textsubscript{$E$}) that maps $X$ to a latent space $z$, a Decoder (G\textsubscript{$D$}) that reconstructs $\hat{X}$ from $z$, and a second Encoder (G\textsubscript{$\hat{E}$}) that maps $\hat{X}$ to $\hat{z}$. The autoencoder corresponds to G\textsubscript{$E$}+G\textsubscript{$D$}. The Discriminator (C) is a convolutional encoder followed by a dense layer that performs binary real/fake classification.
					
					The generator objective is composed of three terms: adversarial, contextual, and encoder-consistency losses.
					
					The adversarial term $\mathcal{L}_{adv}$ stabilizes optimization by matching feature activations from C. Let $\mathcal{F}_C$ denote the output of the last convolutional layer of C:
				\begin{equation}
					\mathcal{L}_{adv} = \mathcal{L}_{2}(\mathcal{F}_C(X), \mathcal{F}_C(\hat{X})),
				\end{equation}
					where $\mathcal{L}_2$ is the $\ell_2$ loss.
					
					The contextual (reconstruction) loss combines pixel-level and structural terms:
				\begin{equation}
					\mathcal{L}_{con} = w_a \cdot \mathcal{L}_{1}(X, \hat{X}) + w_b \cdot \mathcal{L}_{\mathrm{SSIM}}(X, \hat{X}),
				\end{equation}
					where $\mathcal{L}_{1}$ is the $\ell_1$ loss and $\mathcal{L}_{\mathrm{SSIM}} = 1 - \mathrm{SSIM}(X, \hat{X})$.
					
					The encoder-consistency loss, as in Akcay et al. \cite{akcay2018ganomaly}, minimizes the distance between $z$ from $X$ and $\hat{z}$ from $\hat{X}$:
				\begin{equation}
					\mathcal{L}_{enc} = \mathcal{L}_{1}(z, \hat{z}).
				\end{equation}
				
					The final generator objective is:
					\begin{equation}
						\mathcal{L}_{gen} = w_1 \cdot \mathcal{L}_{adv} + w_2 \cdot \mathcal{L}_{con} + w_3 \cdot \mathcal{L}_{enc},
					\end{equation}
					where $w_a$ and $w_b$ weight the contextual subterms, and $w_1$, $w_2$, $w_3$ weight the three components of $\mathcal{L}_{gen}$.
				}
				
			\subsection{Application specific optimizations}
					\label{sec: optimizations}
				{	
					Starting from the framework described in section \ref{sec: grd-net}, we introduced application-specific augmentations: Perlin-noise perturbation, random rotation in $[- \frac{\pi}{8}, \frac{\pi}{8}]$ rad, and random vertical flip. Horizontal flips and rotations larger than $\frac{\pi}{2}$ were excluded because they can generate unrealistic anomalous patterns.
					
					These augmentations improve generalization but also increase training entropy, which can reduce optimization stability. To counter this effect, we introduced a \textit{Noise Loss}:
				\begin{equation}
					\mathcal{L}_{nse} = w_4 \cdot \mathcal{L}_{2}(|(1.0 - \beta) {\hat{M} \cdot \hat{X}} - {M \cdot \overstar{X}}|, \beta N),
				\end{equation}
				
					where \(\overstar{X}\) denotes the post-perturbation input generated by \(P_{q}(X)\):
				
				\begin{equation}
					\begin{split}
						\overstar{X} &= (1 - M) \cdot X + (1.0 - \beta) M \cdot X + \beta N, \\
						\beta &\sim \mathcal{U}(0.5, 1.0), \\
						N &= \text{Perlin noise isolated}, \\
						M &= 
						\begin{cases}
							1, & \text{if the pixel belongs to the noise region } N, \\
							0, & \text{otherwise}
						\end{cases}
					\end{split}
					\label{eq:perlin_noise_function}
				\end{equation}
					\begingroup
					\renewcommand{\mathbb}[1]{\text{\usefont{U}{bbold}{m}{n}#1}}
					where \(\overstar{X}, M, N, \beta = P_{q}(X)\), and \(N\) is the isolated Perlin-noise map applied to a black image with the same size as \(X\). Equation \ref{eq:perlin_noise_function} defines the superimposition mechanism. Parameter \(q\) is the probability of applying Perlin noise; therefore, with probability \(1-q\), \(\overstar{X} = X\), \(M = \mathbb{0}\), and \(N = \mathbb{0}\). Perlin noise introduces non-Gaussian perturbations that better resemble real defects and simultaneously masks non-rectangular regions \cite{DBLP:journals/corr/abs-2111-06377,prabhakar2023vitaeimprovingvisiontransformer}. This discourages trivial identity mapping, a common weakness of vanilla autoencoders on small defects.
					\endgroup
					Moreover, the $\ell_1$ term in the contextual loss was replaced with Huber loss to improve stability around the origin:
					\begin{equation}
						\mathcal{L}_{con} = w_a \cdot \mathcal{L}_{Huber}(X, \hat{X}) + w_b \cdot \mathcal{L}_{\mathrm{SSIM}}(X, \hat{X}),
					\end{equation}
					
					Starting from DR\AE M \cite{zavrtanik2021draem} and GANomaly \cite{akcay2018ganomaly}, we used an iterative branch-and-bound tuning process and obtained the following configuration:
				\begin{equation}
				\begin{split}
					w_a & = 2.0  \\
					w_b & = 1.0  \\
					w_1 & = 1.0  \\
					w_2 & = 50.0 \\
					w_3 & = 1.0  \\
					w_4 & = 3.0,
				\end{split}
				\end{equation}
					Although $\mathcal{L}_{\mathrm{SSIM}}$ provides the strongest contribution to reconstruction quality, it can be unstable on high-entropy images. Increasing $w_a$ mitigated this behavior at the cost of slower convergence.
				}
			
			\subsubsection{Network}
			\label{sec: network}
			{
				The network used as generator for training is an encoder-decoder-encoder shaped net, where the encoder has a ResNet architecture (Figure \ref{fig:enc_net}), and the decoder (Figure \ref{fig:dec_net}) a reverse residual shape. As mentioned in section \ref{sec: autoencoders}, residual network are widely used \cite{he2015deep,9373350,Zini_2020,Le2023}, since they prevent several degradation problems during training, as the gradient vanishing. In the context of this work, residual structure was updated to the most recent ones, using as starting point the one designed in the previous work \cite{GRDNetArticle}.
				
				\begin{figure}
					\centering
					\subfloat[Residual encoder]{\includegraphics[scale=0.55]{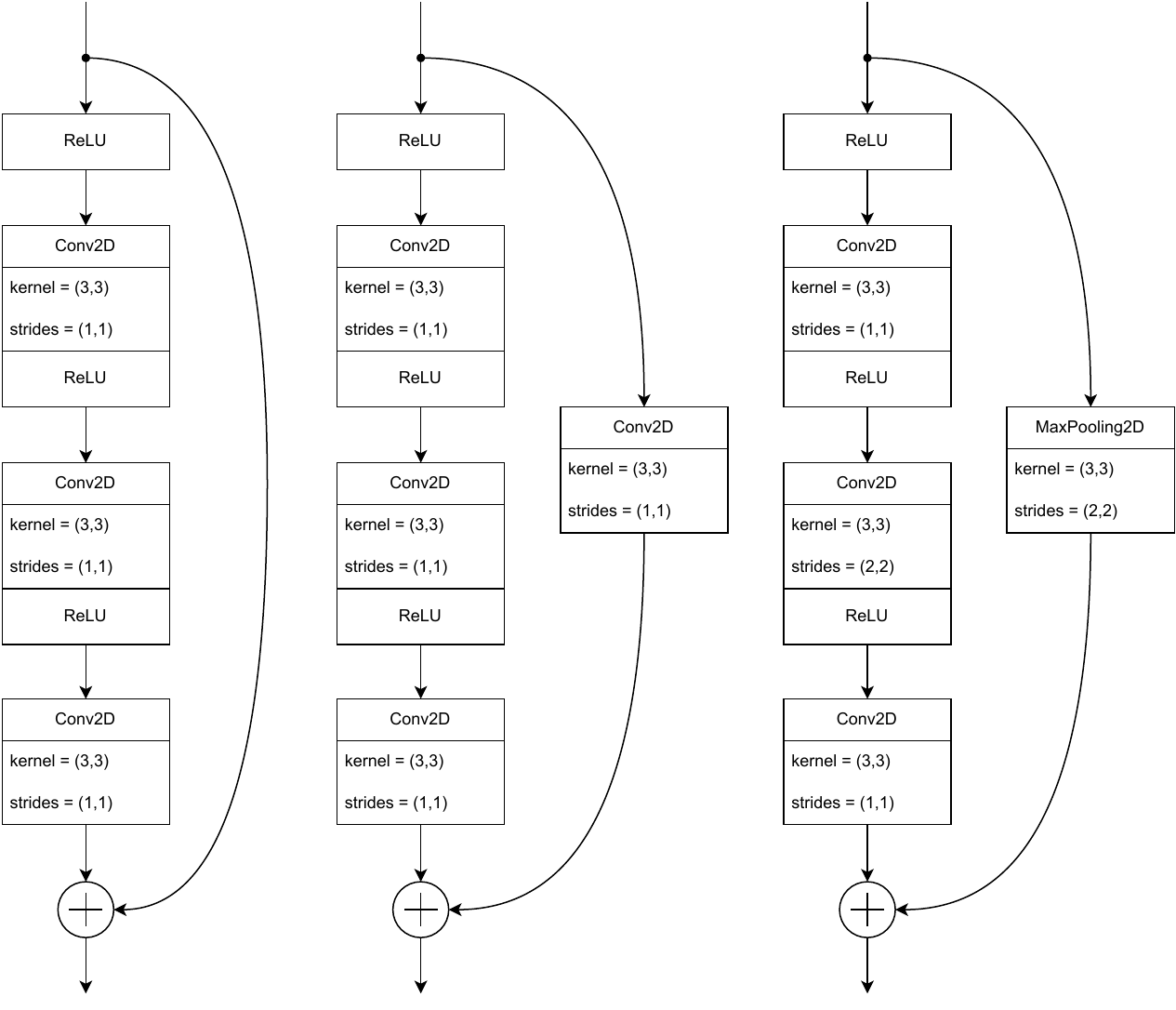}}
					\caption{Three residual blocks in the encoder network. Only the last one halve the $(H,W)$ size of the layer}
					\label{fig:enc_net}
				\end{figure}
				\begin{figure}
				\centering
				\subfloat[Residual decoder]{\includegraphics[scale=0.55]{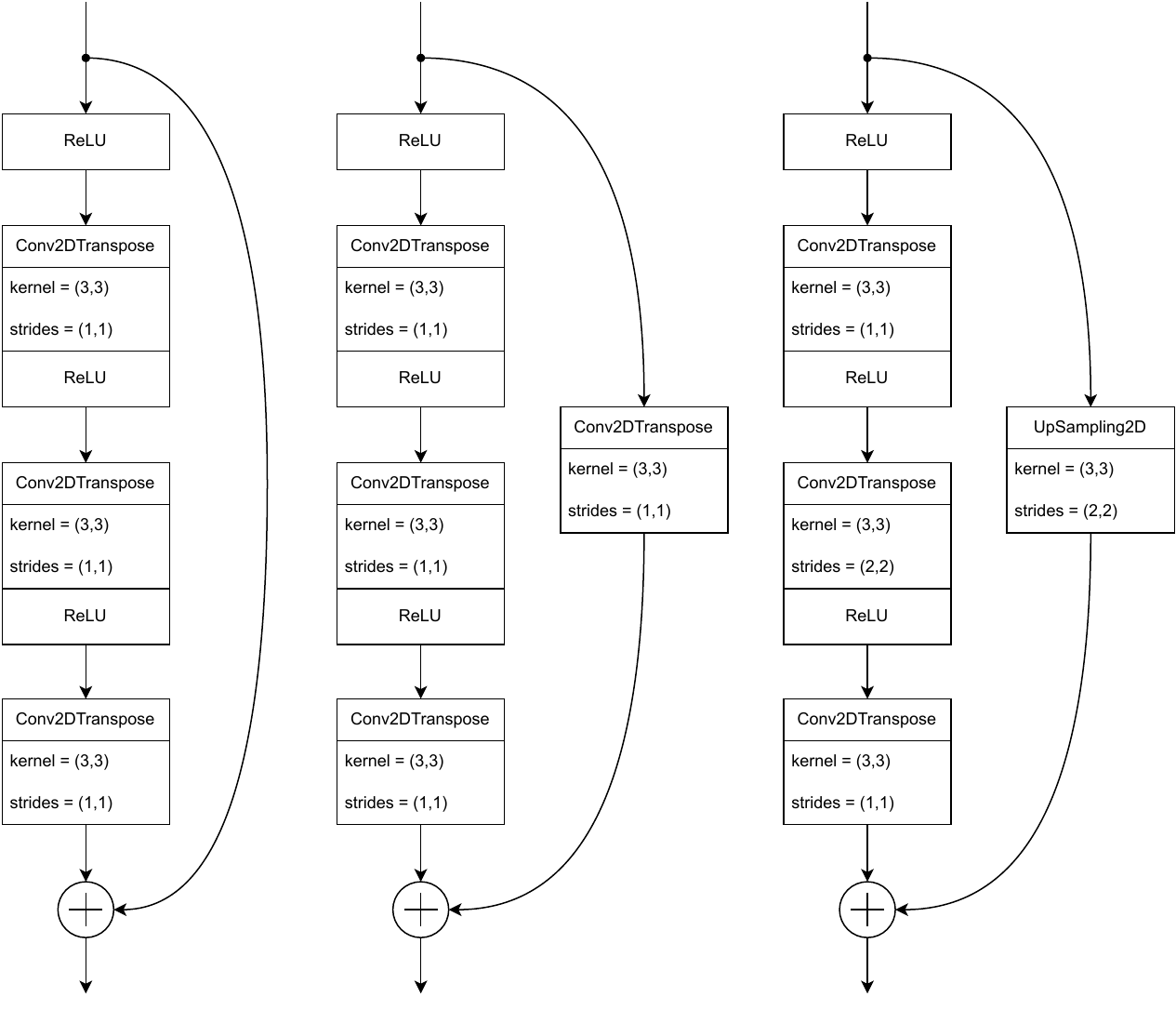}}
				\caption{Three residual blocks in the decoder network. Only the last one double the $(H,W)$ size of the layer}
				\label{fig:dec_net}
				\end{figure}
			}
			
			\subsubsection{Classification and segmentation}
			\label{sec: classification_segmentation}
			{
					This application does not require full-resolution semantic segmentation; patch-level anomaly classification and heatmap localization are sufficient for process integration. Therefore, only the generative branch is used in training and inference. Patch partitioning allows coarse spatial localization of anomalies.
					The anomaly score is defined as:
				\begin{equation}
					\phi = 1 - \mathrm{SSIM}(X, \hat{X}),
					\label{eq:anomaly_score}
				\end{equation}
					and the heatmap is computed as the absolute difference between input and reconstruction:
				\begin{equation}
					H = |X - \hat{X}| \bigg|^1_0,
				\label{eq: heatmap_eq}
				\end{equation}
					where $\bigg|^1_0$ denotes min-max normalization in $[0,1]$:
					\begin{equation}
						H = a + b \cdot \frac{|X - \hat{X}| - \mathrm{min}(|X - \hat{X}|)}{\mathrm{max}(|X - \hat{X}|) - \mathrm{min}(|X - \hat{X}|)}, ~ a = 0 \wedge b = 1.
						\label{eq: min-max-norm}
					\end{equation}
				
					The optimal threshold $\phi_t$ is tuned on a dedicated calibration subset containing both nominal and defective samples, strictly disjoint from the final test kit used for reporting. During inference, if one patch is classified as reject, the corresponding product is flagged as reject. In that case, \ref{eq: heatmap_eq} is used to generate the heatmap of the anomalous subregion.
				}
		}
	}

	\section{Experiments}
		\label{sec: experiments}
		{
			We conducted a set of experiments to optimize detection performance while preserving inference latency within the 500 ms slot between two consecutive acquisitions. For completeness, we first summarize the industrial context and the hardware/software environment.
		
			\paragraph{Product}
			{
				The inspected product (Figure \ref{fig:product}\footnote{\label{foot:nda}Because of an NDA, the \textit{flag} (upper part) was blurred to conceal the company logo.}) is a BFS strip of 5 vials filled with 10 ml of liquid excipient. During handling, bubbles are frequently generated in the liquid region, especially around the \textit{meniscus}. In addition, the narrow \textit{neck} can trap liquid, and droplets are often visible above the liquid level.
				As shown in Figure \ref{fig:regions}, each strip contains 5 vials, and each vial is partitioned into 4 logical regions: \textit{flag} (upper, red), top \textit{body} (blue), liquid \textit{body} (green), and \textit{bottom} (yellow).
			\begin{figure}[htp]
				\centering
				\subfloat[The product]{\includegraphics[scale=0.55]{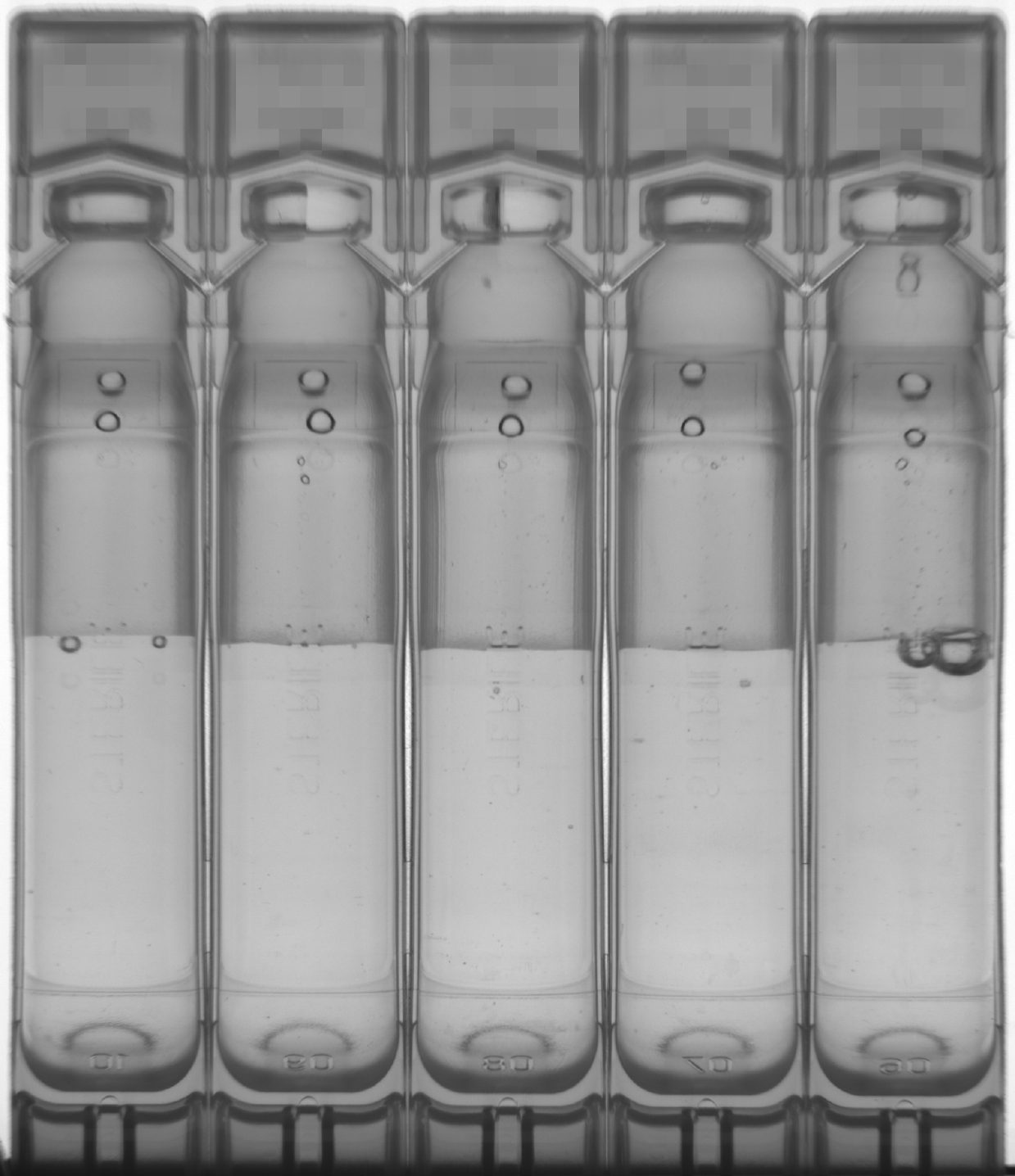}}
				\caption[]{The product: a BFS strip composed by 5 vials\textsuperscript{\ref{foot:nda}}.}
				\label{fig:product}
			\end{figure}
			\begin{figure}
				\centering
				\subfloat[The vial regions]{\includegraphics[scale=0.55]{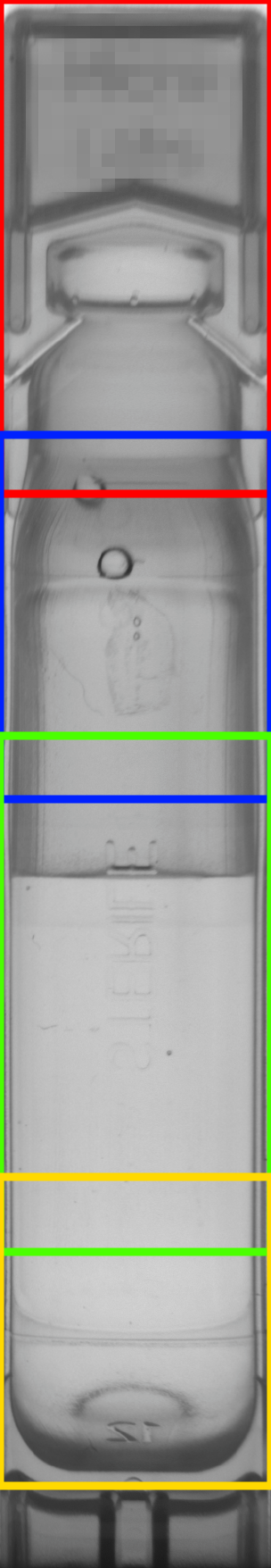}}
				\caption[]{The vial regions\textsuperscript{\ref{foot:nda}}: the logic regions in which the vial is divided.}
				\label{fig:regions}
			\end{figure}
		}
		
			\paragraph{Train dataset acquisition}
			{
				The production acquisition setup was replicated at laboratory scale. Both setups used a telecentric lens to mitigate geometric distortion, especially on side and bottom vial regions. Each strip was acquired 10 times with a semicircular motion focused on the upper side of the product, yielding 16 frames per acquisition. For each frame set, two additional images were generated via a \textit{rank} filter by selecting the minimum and maximum gray-level responses.
				From MVtec definition of the filter\footnote{\url{https://www.mvtec.com/doc/halcon/2305/en/rank_image.html}}:
				Conceptually, the rank filter sorts all gray values within the mask in ascending order and then selects the gray value with rank $Rank$. The rank $1$ corresponds to the smallest gray value and the rank $A$\footnote{In this case $A = 16$} corresponds to the largest gray value within the mask.
				As a consequence, moving artifacts are suppressed in lighter-rank images and emphasized in darker-rank images. This increases the variability of nominal samples.
				After patch extraction, the dataset contained 2,815,200 images, split into 90\% training and 10\% validation.
				Each patch is a grayscale image of size $256 \times 256 \times 1$.
			}
		
		\paragraph{Test dataset acquisition} {
			A real-case set of defective strips was acquired in order to benchmark the algorithm's performance. Each acquisition is made of three frames, which are the first, the eighth, and the last in the series, respectively, to lessen the computational load on the system while it is operating. Consequently, the numerosity ($N$) of the batch is:
			\begin{equation}
				N = f \cdot v \cdot r = 3 \cdot 5 \cdot 4 = 60,
			\end{equation}
			where $f$ is the frame number, $v$ is the vial number and finally, $r$ is the region number.
		}
		
		\paragraph{Machine handling} {
			The machine is a rotating inline quality control machine that inspects the products at the end of the production line. Usually these machines are placed right after the filling device and before the packaging ones.
			
			The product is gripped from the input conveyor belt and brought into the carousel and handled identically to how it was handled during the laboratory acquisitions. The handling and movement inside the carousel is the same as the one in the laboratory mockup, but that there is additional movement of the liquid inside due to how the product is handled before the carousel, but it has been observed on previous machines, that this further movement does not affect the performance of the algorithm, since it is much less and far away in time than the one produced on the inspection station.
			After handling the product, acquisitions take place, and the images begin to be processed\footnote{More specific details about machine hardware cannot be provided due to the NDA and company policies}.
		}

			\medskip
			\paragraph{Experiments} { Training was run for 10 epochs because of the very high numerosity of the training set, using the pipeline described in section \ref{sec: methods}, and specifically in subsections \ref{sec: optimizations}, \ref{sec: network}, and \ref{sec: classification_segmentation}. The residual backbone follows the ResNet v2 design principles \cite{he2015deep}.
			This section reports:
			\begin{enumerate}
				\item The hardware and software environment
				\item The architecture of the adopted network
				\item The training phase
				\item Performance on positive and negative examples in the test dataset
				\item The results in terms of time and throughput
				\item Experiments on the generative network
			\end{enumerate}
			}
		
		\subsection{Hardware} {
			The hardware, as already mentioned in subsection \ref{sec: sol-work} is very different between the training server and the inference machine. The training process is carried out on a server that has been set up in this way:
			\begin{itemize}
				\item Hardware
				\begin{itemize}
					\item Intel\textsuperscript{\textregistered} Xeon\textsuperscript{\textregistered} Silver 4216 CPU @ 2.10 GHz as CPU
					\item 64GB of DDR4 Synchronous @ 3200 MHz as system memory
					\item Nvidia\textsuperscript{\textregistered} A100 with 40GB of VRAM as GPU.
					\item 2 TB M.2 NVMe SSD
				\end{itemize}
				\item Software
				\begin{itemize}
				\item Ubuntu 22.04.3 LTS Server minimal-based o/s with 5.15.0-94-x86\_64 kernel
				\item Nvidia\textsuperscript{\textregistered} driver Version: 535.104.12
				\item CUDA\textsuperscript{\textregistered} version: 12.2
				\item TensorFlow 2.13.1 compiled from source
				\end{itemize}
			\end{itemize}
			The inference machine is an industrial cluster installed on board and it is set up in this way:
			\begin{itemize}
				\item Hardware
				\begin{itemize}
					\item Intel\textsuperscript{\textregistered} Xeon\textsuperscript{\textregistered} E-2278GE CPU @ 3.30 GHz as CPU
					\item 32GB of DDR4 Synchronous @ 3200 MHz as system memory
					\item Nvidia\textsuperscript{\textregistered} A4500 with 20GB of VRAM as GPU.
					\item 32 GB CFast flash for the operative system (read-only mounted)
					\item 1 TB M.2 NVMe SSD for the data
				\end{itemize}
				\filbreak
				\item Software
				\begin{itemize}
					\item Ubuntu 22.04.3 LTS Server minimal-based o/s with 5.15.0-94-x86\_64 kernel
					\item Nvidia\textsuperscript{\textregistered} driver Version: 535.104.12
					\item CUDA\textsuperscript{\textregistered} version: 12.2
					\item TensorFlow 2.13.1 compiled from source
				\end{itemize}
			\end{itemize}
		}
		
		\subsection{Network architecture} {
			\paragraph{Encoder} {
				As mentioned in subsection \ref{sec: network}, the network is residual based on ResNet version 2 architecture.
				\begin{figure}
					\centering
					\subfloat[ResEnc Black A]{\includegraphics[scale=0.55]{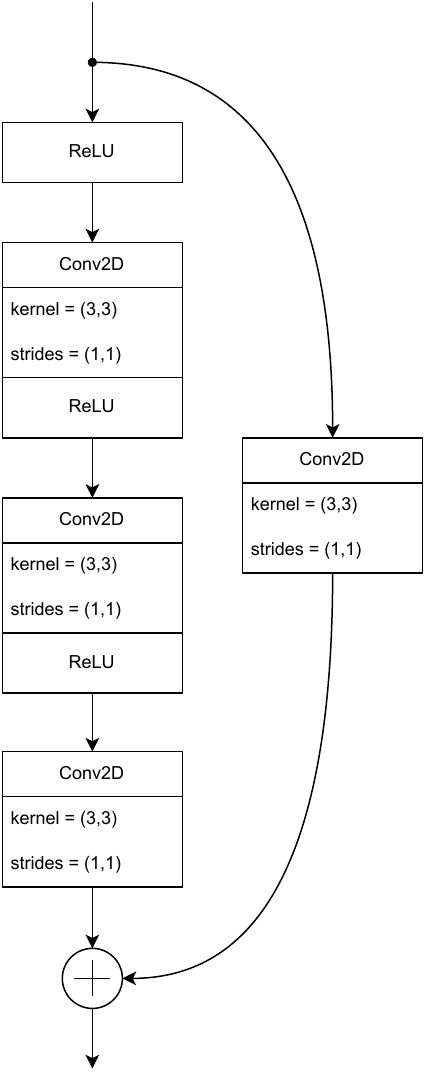}}
					\hspace{0.25cm}
					\subfloat[ResEnc Block B]{\includegraphics[scale=0.55]{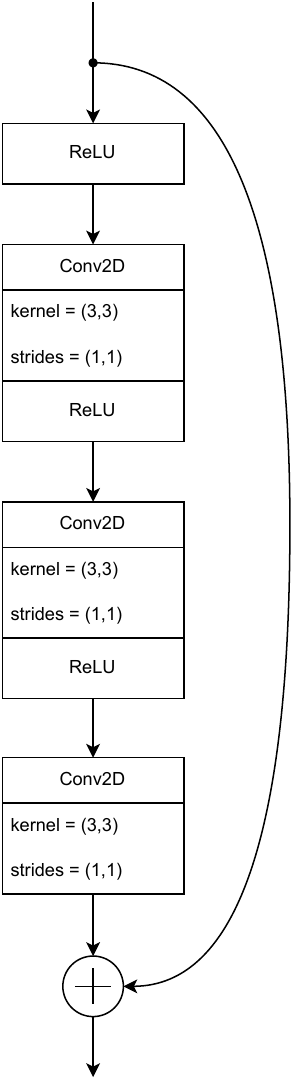}}
					\hspace{0.25cm}
					\subfloat[ResEnc Block C]{\includegraphics[scale=0.55]{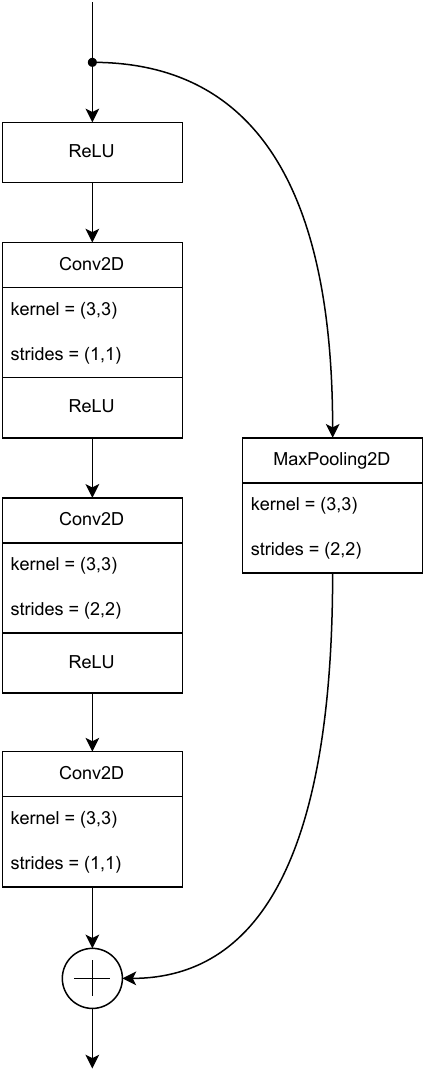}}
					\caption{Three residual blocks in the encoder (ResEnc). One stage consists of three convolutional blocks in the order A, B, C.}
					\label{fig:enc_stage_blocks}
				\end{figure}
				Each stage is composed of 3 residual blocks:
				\begin{enumerate}
					\item The first block (A in Figure \ref{fig:enc_stage_blocks}) concatenates three consecutive $3 \times 3$ convolutions with one $1 \times 1$ convolution. Spatial size $H_i \times W_i$ is preserved.
					\item The second block (B in Figure \ref{fig:enc_stage_blocks}) concatenates three consecutive $3 \times 3$ convolutions with the input from block A. Spatial size $H_i \times W_i$ is preserved.
					\item The third block (C in Figure \ref{fig:enc_stage_blocks}) concatenates three consecutive $3 \times 3$ convolutions with a downsampling branch. The middle convolution halves both $H_i$ and $W_i$.
				\end{enumerate}
				The encoder uses 4 stages, yielding a final embedding size of $16 \times 16$ with 1024 channels.
			}
			\paragraph{Bottleneck} {
				The bottleneck is fully-connected with a size of $64$ features, as in figure \ref{fig:bottleneck}.
				\begin{figure}
					\centering
					\subfloat[Dense bottleneck]{\includegraphics[scale=0.55]{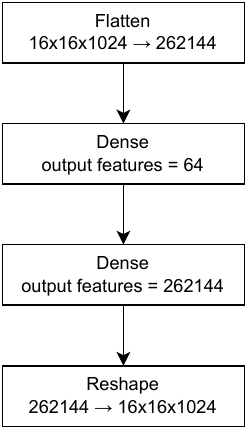}}
					\caption{The bottleneck stage.}
					\label{fig:bottleneck}
				\end{figure}
			}
			\paragraph{Decoder} {
				The decoder is the inverse architecture of the encoder, described above.
				\begin{figure}
					\centering
					\subfloat[ResDec Black A]{\includegraphics[scale=0.55]{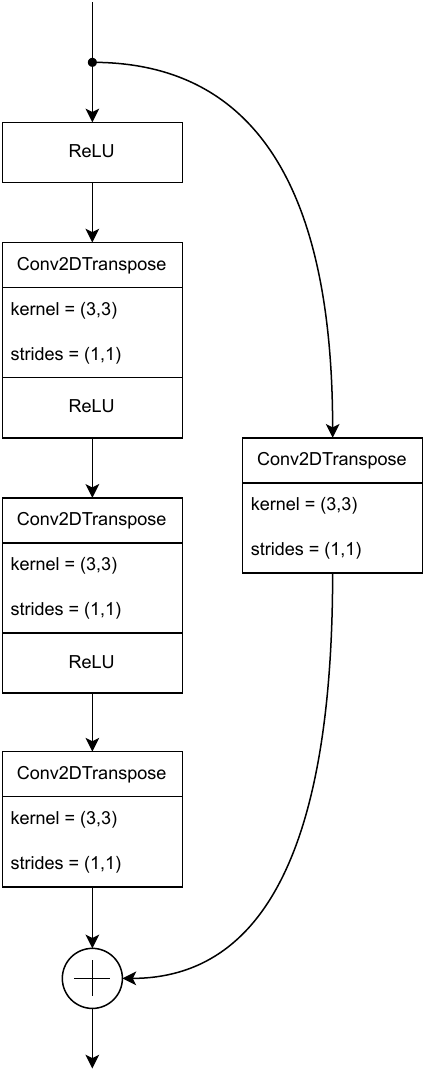}}
					\hspace{0.25cm}
					\subfloat[ResDec Block B]{\includegraphics[scale=0.55]{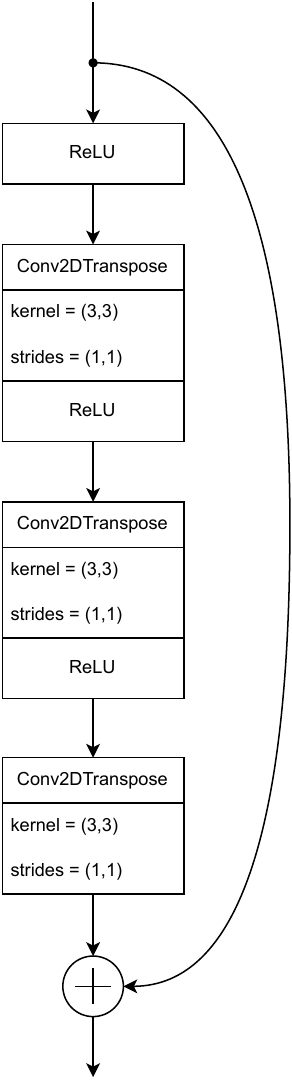}}
					\hspace{0.25cm}
					\subfloat[ResDec Block C]{\includegraphics[scale=0.55]{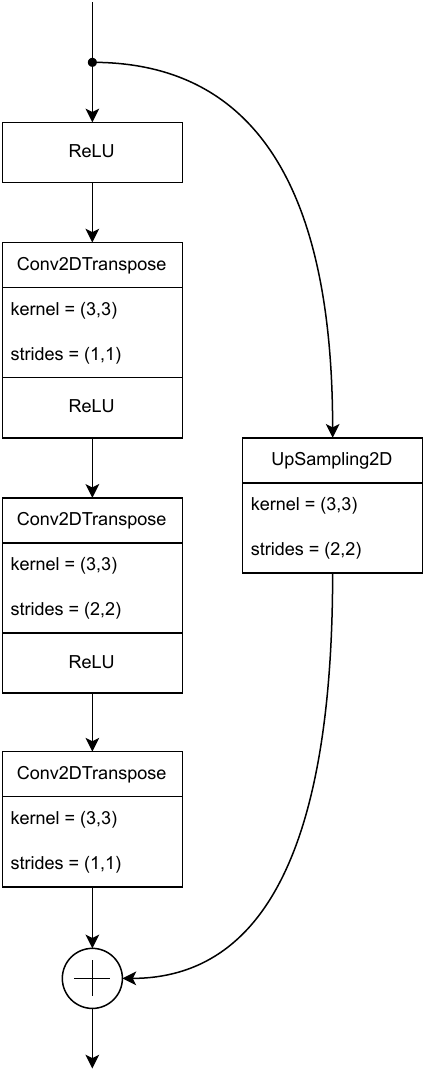}}
					\caption{Three residual blocks in the decoder (ResDec). One stage consists of three transposed-convolution blocks in the order A, B, C.}
					\label{fig:dec_stage_blocks}
				\end{figure}
				As in the encoder, each stage is composed of 3 residual blocks:
				\begin{enumerate}
					\item The first block (A in Figure \ref{fig:dec_stage_blocks}) concatenates three consecutive transposed convolutions with $3 \times 3$ kernel and one $1 \times 1$ convolution. Spatial size $H_i \times W_i$ is preserved.
					\item The second block (B in Figure \ref{fig:dec_stage_blocks}) concatenates three consecutive transposed convolutions with the input from block A. Spatial size $H_i \times W_i$ is preserved.
					\item The third block (C in Figure \ref{fig:dec_stage_blocks}) concatenates three consecutive transposed convolutions with an upsampling branch. The middle convolution doubles both $H_i$ and $W_i$.
				\end{enumerate}
				The decoder uses 4 stages to recover a final output size of $256 \times 256$ with a single channel and sigmoid activation.
			}
			}
		
		\clearpage
		\subsection{Training phase} {
			Training follows a standard GAN schedule with alternating optimization of generator and discriminator. The initial learning rate is $1.5 \times 10^{-4}$, with cosine-decay restarts every $2{,}533{,}680$ steps and restart amplitude set to $\frac{1}{3}$ of the previous maximum. Adam is used for both networks. Batch size is 32, and perturbation probability is set to $q = 0.75$.
			\begin{figure}
				\centering
				\subfloat[Train flowchart]{\includegraphics[scale=0.55]{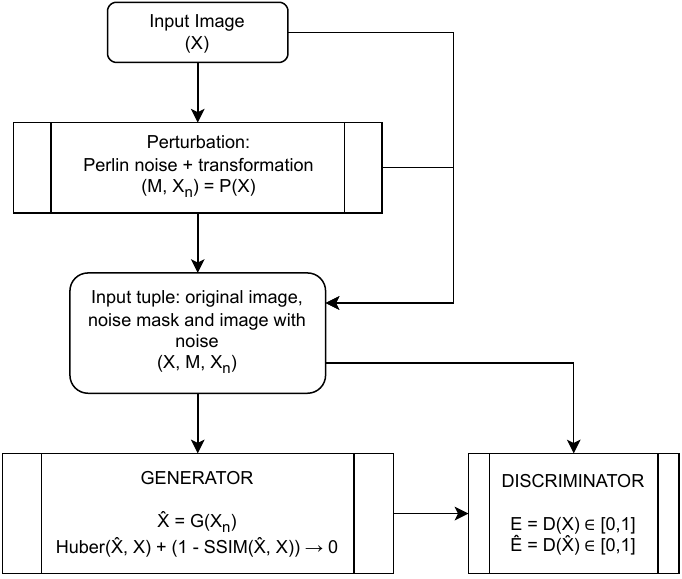}}
				\caption{The training flowchart.}
				\label{fig:training-flowchart}
			\end{figure}
			The training flowchart is shown in Figure \ref{fig:training-flowchart}.
		}
		
			\subsection{Results} {
				Experiments were performed on the client-provided knapp test kit. The set contains 141 defective products and 120 nominal products. Since the class distribution is relatively balanced, we report accuracy, true positive ratio (TPR), true negative ratio (TNR), balanced accuracy, and inference time statistics. Thresholds are calibrated on a dedicated validation subset, and the final test kit is used only for performance reporting. In this manuscript, we report point estimates; confidence intervals will be included in an extended statistical analysis.
				The analysis is presented at three aggregation levels:
				\begin{enumerate}
					\item Overall accuracy on the whole test dataset and inference time per patch
					\item Accuracy after image aggregation per product and inference time per product
					\item Accuracy, after image aggregation per product and run, based on acceptance policy of the client
				\end{enumerate}
				A representative qualitative subset is reported in Appendix \ref{apx:res_images}.
				
				Due to NDA constraints on the proprietary dataset and pipeline, this manuscript focuses on absolute deployment performance and engineering feasibility. A full baseline comparison against representative methods (e.g., PaDiM, PatchCore, FastFlow) on public benchmarks with the same preprocessing and evaluation protocol is planned as follow-up work.
			
				\paragraph{Acceptance policy} {
					For industrial validation, the machine must perform at least as well as human inspectors on the same knapp test kit. Each product is acquired 10 times; each acquisition is defined as one run.
					A nominal product is considered correctly classified if it is predicted as nominal in at least 7 runs out of 10. A defective product is considered correctly classified if it is predicted as defective in at least 7 runs out of 10. Otherwise, it is counted as misclassified.
				}
			
				\paragraph{Overall results} {
					Overall results are reported in table \ref{tab:overall_res}. A dedicated threshold is estimated for each region due to region-specific pixel distributions.
					The mean inference time per frame is computed as $\mu_{t_{f}} = \frac{\mu_{t_{b}}}{f \cdot v \cdot r} = \frac{\mu_{t_{b}}}{60}$, where $\mu_{t_{f}}$ is the mean time per frame, $\mu_{t_{b}}$ is the mean time per batch, $v$ is the number of vials per strip, and $r$ is the number of regions per vial.
				
				\begin{table}[htbp]
						\centering
						\begin{tabular}{llllll}
							\hline
							\multicolumn{6}{c}{\textbf{Overall results}}                                                           \\ \hline
							\multicolumn{1}{l|}{}                                  & \textbf{R0} & \textbf{R1}  & \textbf{R2} & \textbf{R3} & \textbf{Vial} \\ \hline
							\multicolumn{1}{l|}{\textbf{Threshold}}                & 0.015589    & 0.038017     & 0.046568    & 0.029593    & -             \\ \hline
							\multicolumn{1}{l|}{\textbf{Accuracy}}                 & 0.9919      & 0.9926       & 0.9957      & 0.9991      & 0.9870  		\\ \hline
							\multicolumn{1}{l|}{\textbf{True positive ratio}}      & 0.9966      & 0.9985       & 0.9986      & 0.9994      & 0.9942  		\\ \hline
							\multicolumn{1}{l|}{\textbf{True negative ratio}}      & 0.9093      & 0.9044       & 0.9779      & 0.9973      & 0.9581  		\\ \hline
							\multicolumn{1}{l|}{\textbf{Balanced accuracy}}        & 0.9530      & 0.9515       & 0.9883      & 0.9984      & 0.9762  		\\ \hline
							\multicolumn{1}{l|}{\textbf{Mean inference time} (ms)} & 0.1689      & 0.1689       & 0.1689      & 0.1689      & -             \\ \hline
							\end{tabular}
						\caption{Overall results using the equation \ref{eq:anomaly_score} as anomaly score.}
						\label{tab:overall_res}
				\end{table}
			}
		
			\paragraph{Per product aggregation} {
				Whole-strip classification is designed to minimize GMP\footnote{Good Manufacturing Practice (GMP) describes the minimum standard that a medicines manufacturer must meet in their production processes.} risk: if at least one region is classified as reject, the entire product is rejected. This policy reduces false acceptance at the expense of increased false rejection; therefore, threshold retuning is required to satisfy project constraints.
				To this end the final score function per vial becomes:
				\begin{equation}
					\theta_v = \mathrm{max}(\left\lbrace 1.0 - \mathrm{SSIM}(X_i, \hat{X}_i) \right\rbrace ), ~ i \ge 0 ~ \land ~ i < {{f} \cdot {{v} \cdot {r}}},
					\label{eq:vial_anomaly_score}
				\end{equation}
				where $i$ is the patch index.
				
				\begin{table}[htbp]
					\centering
					\begin{tabular*}{0.75\textwidth}{ll@{\extracolsep{\fill}}}
						\hline
						\multicolumn{2}{c}{\textbf{Per strip results}}                      \\ \hline
						\multicolumn{1}{l|}{}                             & \textbf{Score} \\ \hline
						\multicolumn{1}{l|}{\textbf{Threshold R0}}        & 0.016156       \\ \hline
						\multicolumn{1}{l|}{\textbf{Threshold R1}}        & 0.038584       \\ \hline
						\multicolumn{1}{l|}{\textbf{Threshold R2}}        & 0.046635       \\ \hline
						\multicolumn{1}{l|}{\textbf{Threshold R3}}        & 0.029660       \\ \hline
							\multicolumn{1}{l|}{\textbf{Accuracy}}            & 0.9593         \\ \hline
							\multicolumn{1}{l|}{\textbf{True positive ratio}} & 0.9694         \\ \hline
							\multicolumn{1}{l|}{\textbf{True negative ratio}} & 0.9467         \\ \hline
							\multicolumn{1}{l|}{\textbf{Balanced accuracy}}   & 0.9581         \\ \hline
							\multicolumn{1}{l|}{\textbf{Mean inference time}} & 0.4873         \\ \hline
						\end{tabular*}
					\caption{Per strip results using the equation \ref{eq:vial_anomaly_score} as anomaly score.}
					\label{tab:pervial_res}
				\end{table}
			}
		
			\paragraph{Per run aggregation} {
				For final validation, the algorithm must match or exceed the manual-inspection benchmark measured by the client on the same knapp kit. Each product is acquired 10 times and is considered correctly classified if the label is confirmed in at least 7 runs out of 10 (70\%). Results are reported in table \ref{tab:perrun_res}.
			
			\begin{table}[htbp]
				\centering
				\begin{tabular*}{0.75\textwidth}{ll@{\extracolsep{\fill}}}
					\hline
					\multicolumn{2}{c}{\textbf{Per strip results}}                      \\ \hline
					\multicolumn{1}{l|}{}                             & \textbf{Score} \\ \hline
					\multicolumn{1}{l|}{\textbf{Threshold R0}}        & 0.013222       \\ \hline
					\multicolumn{1}{l|}{\textbf{Threshold R1}}        & 0.034650       \\ \hline
					\multicolumn{1}{l|}{\textbf{Threshold R2}}        & 0.046201       \\ \hline
					\multicolumn{1}{l|}{\textbf{Threshold R3}}        & 0.029226       \\ \hline
						\multicolumn{1}{l|}{\textbf{Accuracy}}            & 0.9641         \\ \hline
						\multicolumn{1}{l|}{\textbf{True positive ratio}} & 0.9676         \\ \hline
						\multicolumn{1}{l|}{\textbf{True negative ratio}} & 0.9599         \\ \hline
						\multicolumn{1}{l|}{\textbf{Balanced accuracy}}   & 0.9638         \\ \hline
					\end{tabular*}
				\caption{Per run results using the equation \ref{eq:vial_anomaly_score} as anomaly score.}
				\label{tab:perrun_res}
			\end{table}
		}
	}

\section{Conclusions and Future work}
	\label{sec: conc_future}
	{
		We developed an architecture that performs efficiently in an extremely demanding industrial environment, where performance has both business and GMP impact, which in turn affects people's safety and health. The proposed method manages a large dataset while adhering to hardware and time limits and striking a fair balance between costs and outstanding performance. In order to identify the primary features of the product and identify out-of-distribution anomalies, it completes the denoising process during training. With the use of this strategy, the network can be forced to perform better than a traditional autoencoder when it comes to compressing and eliminating extraneous features from the result. In fact, because autoencoder-based networks, like GANomaly, have learnt too smoothly to replicate the source image, they frequently have the issue of being able to repeat the anomaly region even in the reconstructed image. Therefore, selecting the noise to be superimposed on the input image becomes crucial since the more similar it is to the distribution of potential anomalies, the higher the yield will be in terms of detecting anomalies. Modern generative models were used as a first instance in the architecture's development, losses and base network design was optimized in order to overcome difficult challenges like the large size of the acquired images and the high variance between positive examples brought on by the movement of the liquid inside the vials. A heatmap for defective items is displayed on the HMI to give the operator a visual explanation of the outcome. The "hot" parts of the heatmap indicate the most anomalous locations in terms of pixels. This architecture was designed jointly by University of Ferrara and Bonfiglioli Engineering, while tests were conducted on the Bonfiglioli Engineering's servers.

		While this research provides insightful information and specifics about how to integrate and implement an out-of-distribution anomaly detection via deep learning algorithm, several areas warrant further exploration. Nowadays, operators frequently need additional explanations in addition to an anomaly score and the associated categorization in order to assess the efficiency of the production line and the machine itself.
		
		Several solutions have been proposed in the embedding similarity-based setting, including PaDiM \cite{defard2021padim}, PatchCore \cite{roth2022towards}, and EfficientAD \cite{batzner2023efficientad}. In these methods, distances between inferred embeddings and nominal embeddings collected during training are projected back to image space to obtain anomaly maps; with calibrated thresholds, this enables pixel-level classification. The reference representation can be organized in multiple forms, such as nominal clusters or probabilistic models.
		Similarly, reconstruction-based networks provide a \textit{heatmap} defined as the absolute value of the difference between the original and the rebuilt images, whereas the differences correspond to the \textit{hottest} parts in the map.
		
		The biggest drawback in both situations is the non-parametric classifier's inability to delve deeper into the result's perceptual interpretation. Further architectures, such as  GRD-Net \cite{GRDNetArticle}, DR\AE M \cite{zavrtanik2021draem}, the described deep perceptual autoencoder by N. Shvetsova et al. \cite{Shvetsova_2021}, or the proposed one by P. Bergmann et al. \cite{Bergmann2022}, have been researched and developed in order to accomplish the aim of identifying defects, within images, interpreting the inferred features. The mentioned architecture prove to have excellent performance, but with the drawback of extreme computational and architectural complexity, which makes them challenging to use in practical situations, particularly when accessed online.
		
		Considering the aforementioned, a promising avenue for future research is to investigate how to obtain information from the latent representation of the input example, as a sample of the stochastic process, that might be processed to produce a visual representation of the anomalous regions within the provided image.
		An excellent starting point can be taken from the work carried out by P. Esser et al. \cite{esser2021taming}, where a discrete representation of the image through a vector quantized variational autoencoder (VQVAE), is used to generate high resolution synthetic images, introducing to this end the use of transformers for imagine encoding, which does not have a strong local correlation within images, in contrast to CNN. However, as the products in this work are the result of an industrial process, their shape and position vary very little. For this reason, a topological connection between the input image and the derived features may help identify anomalies that deviate from the distribution.
	}

\FloatBarrier
\appendix
\section{Result Images}
\label{apx:res_images}{
	Below are shown examples taken from real cases analyzed during algorithm validation phase.
	
	\begin{figure}[H]
		\centering
		\subfloat[Original vial region image ($X$)]{\includegraphics[scale=0.4]{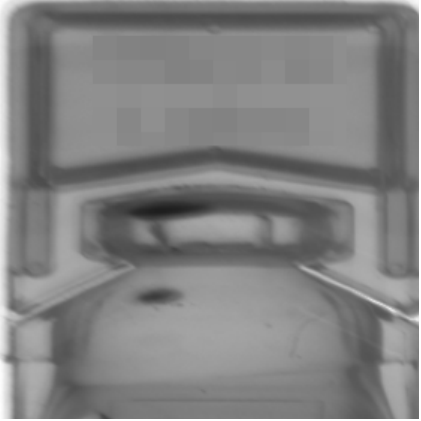}}
		\hspace{0.25cm}\subfloat[The reconstructed image $\hat{X}$]{\includegraphics[scale=0.4]{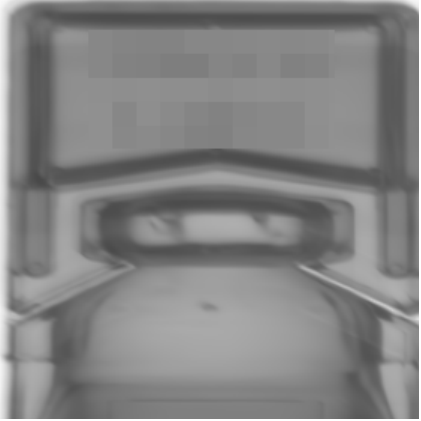}} \\
		\subfloat[The difference image]{\includegraphics[scale=0.4]{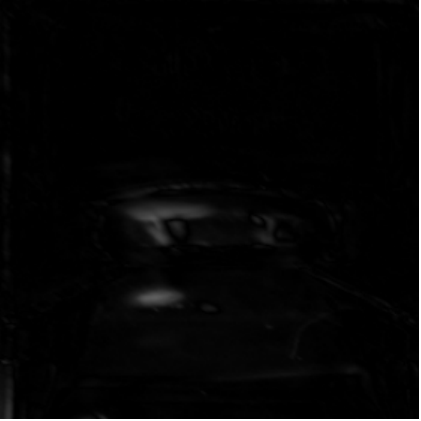}}
		\hspace{0.25cm}\subfloat[The heatmap $H$ superimposed]{\includegraphics[scale=0.4]{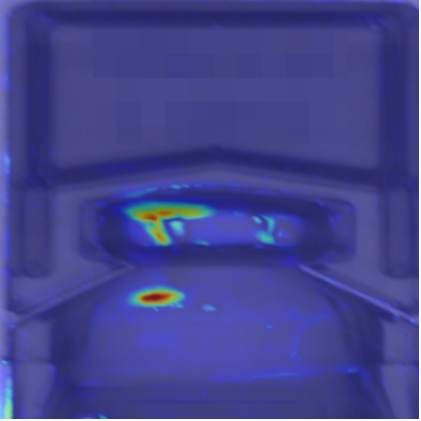}}
		\caption{Stuck particle in the upper part of the product}
		\label{fig:1_ex}
	\end{figure}

	\begin{figure}[H]
		\centering
		\subfloat[Original vial region image ($X$)]{\includegraphics[scale=0.4]{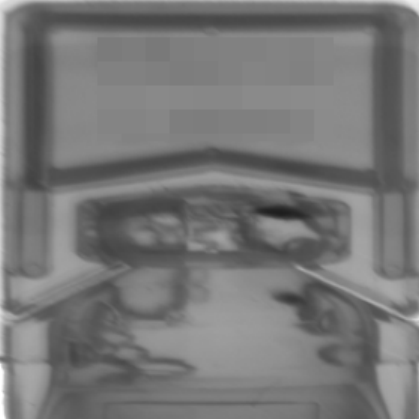}}
		\hspace{0.25cm}\subfloat[The reconstructed image $\hat{X}$]{\includegraphics[scale=0.4]{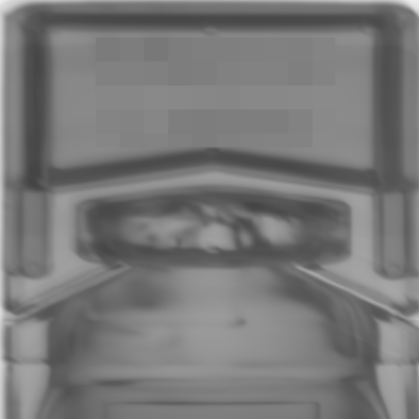}} \\
		\subfloat[The difference image]{\includegraphics[scale=0.4]{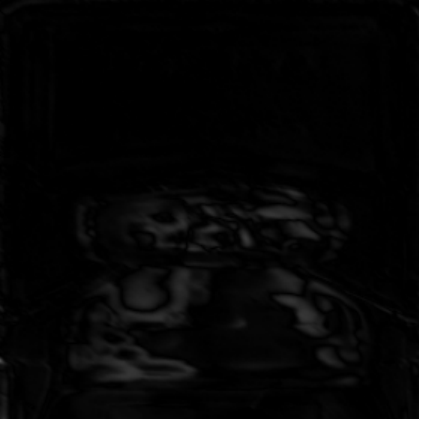}}
		\hspace{0.25cm}\subfloat[The heatmap $H$ superimposed]{\includegraphics[scale=0.4]{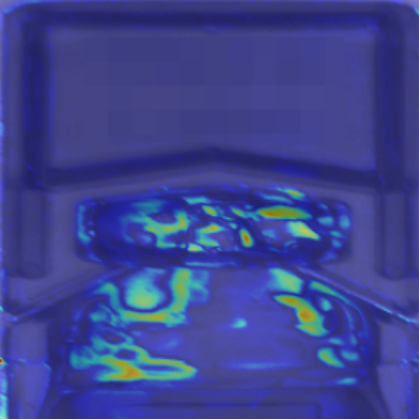}}
		\caption{Stuck particle and anomalous liquid behavior that results in foam formation because of contamination}
		\label{fig:2_ex}
	\end{figure}

	\begin{figure}[H]
		\centering
		\subfloat[Original vial region image ($X$)]{\includegraphics[scale=0.4]{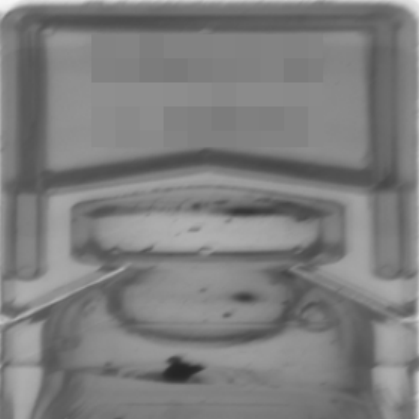}}
		\hspace{0.25cm}\subfloat[The reconstructed image $\hat{X}$]{\includegraphics[scale=0.4]{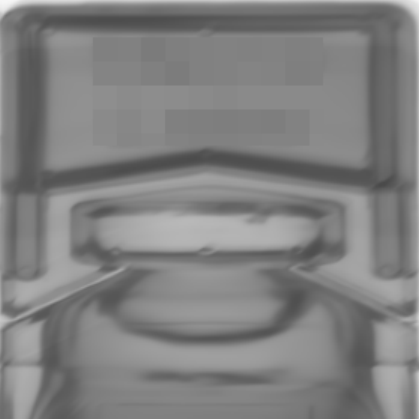}} \\
		\subfloat[The difference image]{\includegraphics[scale=0.4]{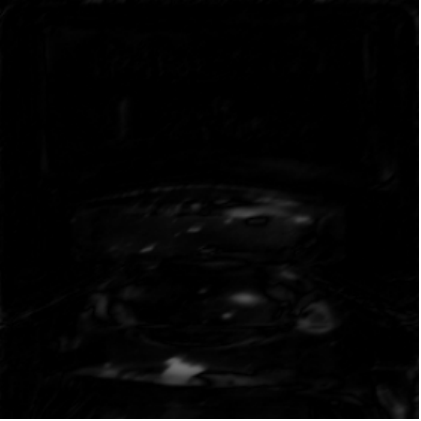}}
		\hspace{0.25cm}\subfloat[The heatmap $H$ superimposed]{\includegraphics[scale=0.4]{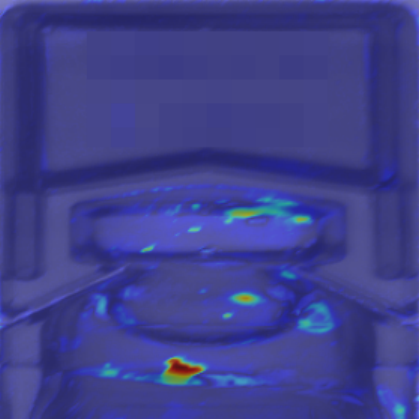}}
		\caption{Stuck particle and anomalous liquid behavior that results in foam formation because of contamination}
		\label{fig:3_ex}
	\end{figure}

	\begin{figure}[H]
		\centering
		\subfloat[Original vial region image ($X$)]{\includegraphics[scale=0.4]{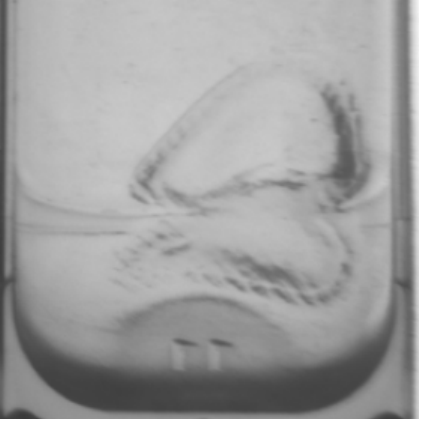}}
		\hspace{0.25cm}\subfloat[The rebuilt image $\hat{X}$]{\includegraphics[scale=0.4]{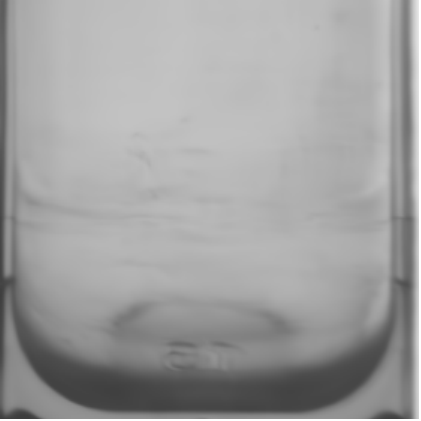}} \\
		\subfloat[The difference image]{\includegraphics[scale=0.4]{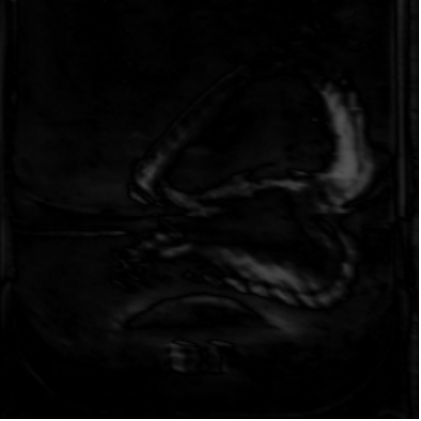}}
		\hspace{0.25cm}\subfloat[The heatmap $H$ superimposed]{\includegraphics[scale=0.4]{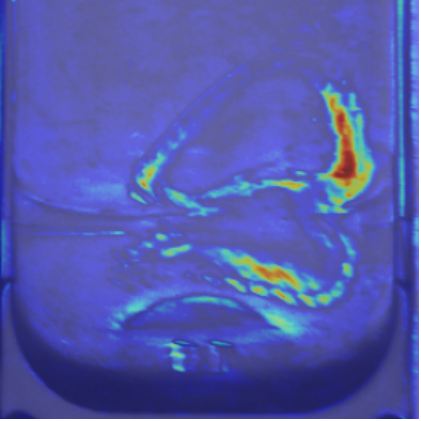}}
		\caption{Lower body deformation}
		\label{fig:4_ex}
	\end{figure}

	\begin{figure}[H]
		\centering
		\subfloat[Original vial region image ($X$)]{\includegraphics[scale=0.4]{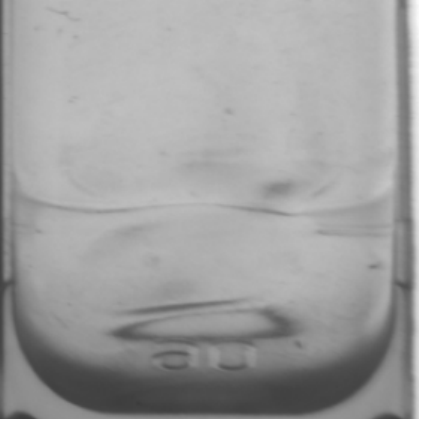}}
		\hspace{0.25cm}\subfloat[The rebuilt image $\hat{X}$]{\includegraphics[scale=0.4]{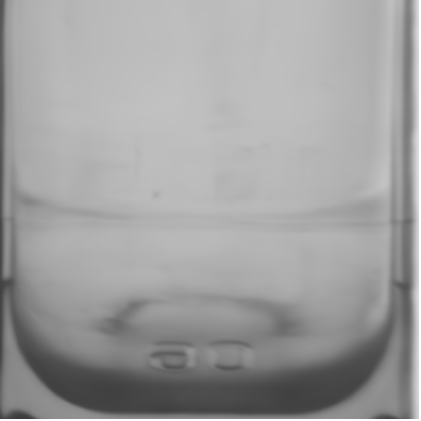}} \\
		\subfloat[The difference image]{\includegraphics[scale=0.4]{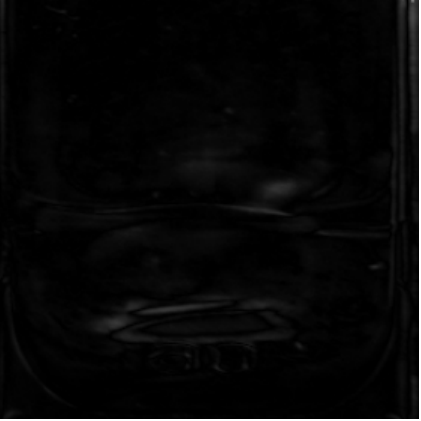}}
		\hspace{0.25cm}\subfloat[The heatmap $H$ superimposed]{\includegraphics[scale=0.4]{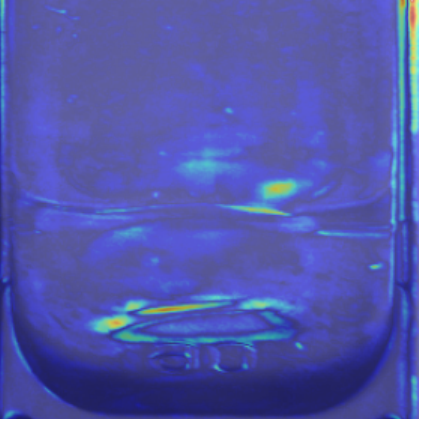}}
		\caption{Light lower body deformation}
		\label{fig:5_ex}
	\end{figure}

	\begin{figure}[H]
		\centering
		\subfloat[Original vial region image ($X$)]{\includegraphics[scale=0.4]{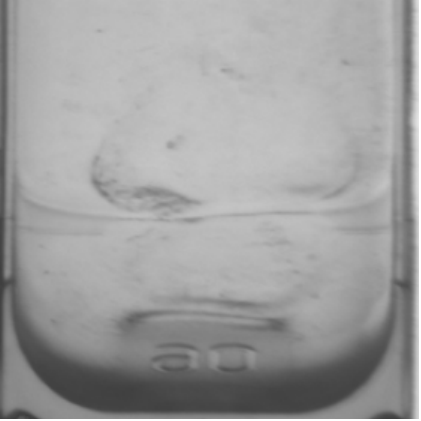}}
		\hspace{0.25cm}\subfloat[The reconstructed image $\hat{X}$]{\includegraphics[scale=0.4]{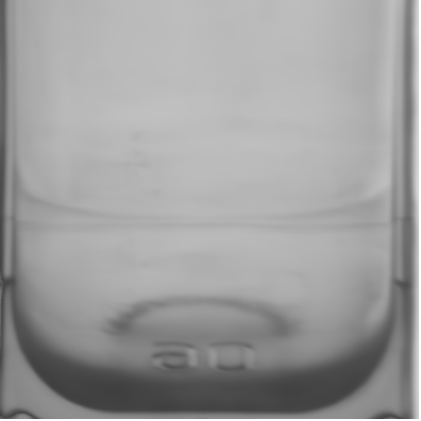}} \\
		\subfloat[The difference image]{\includegraphics[scale=0.4]{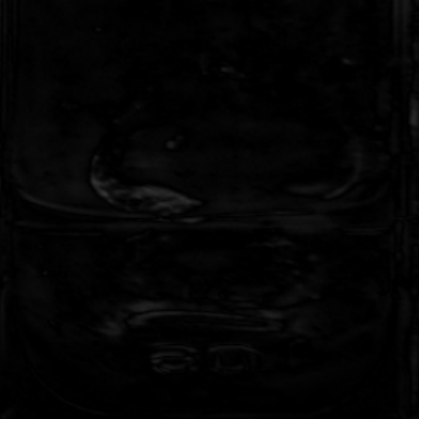}}
		\hspace{0.25cm}\subfloat[The heatmap $H$ superimposed]{\includegraphics[scale=0.4]{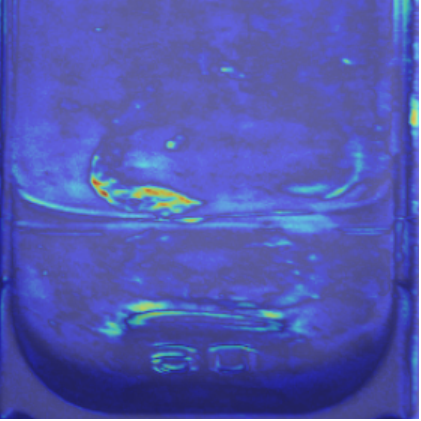}}
		\caption{Lower body deformation}
		\label{fig:6_ex}
	\end{figure}

	\begin{figure}[H]
		\centering
		\subfloat[Original vial region image ($X$)]{\includegraphics[scale=0.4]{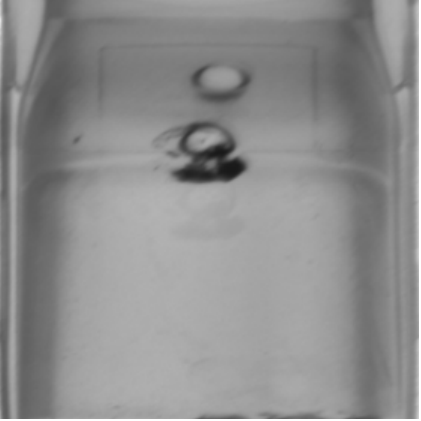}}
		\hspace{0.25cm}\subfloat[The reconstructed image $\hat{X}$]{\includegraphics[scale=0.4]{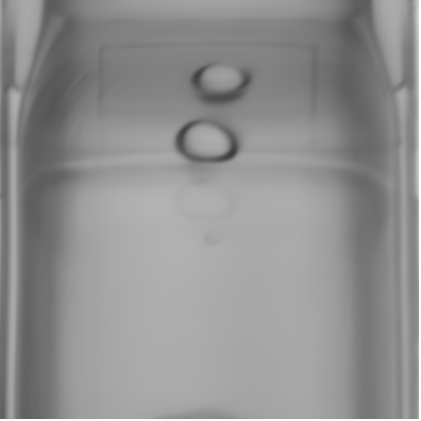}} \\
		\subfloat[The difference image]{\includegraphics[scale=0.4]{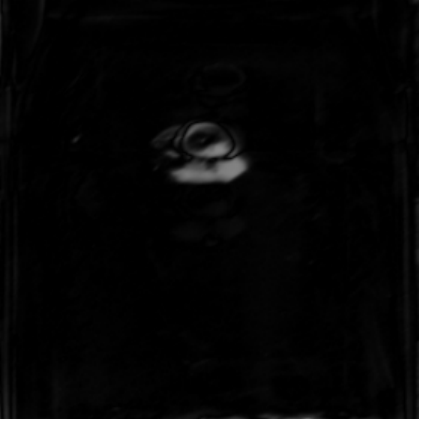}}
		\hspace{0.25cm}\subfloat[The heatmap $H$ superimposed]{\includegraphics[scale=0.4]{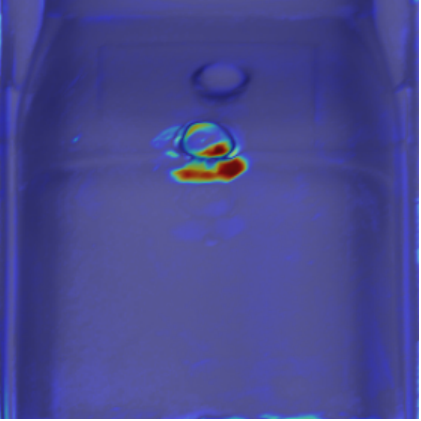}}
		\caption{Black spot on upper part of the vial's body.}
		\label{fig:7_ex}
	\end{figure}

	\begin{figure}[H]
		\centering
		\subfloat[Original vial region image ($X$)]{\includegraphics[scale=0.4]{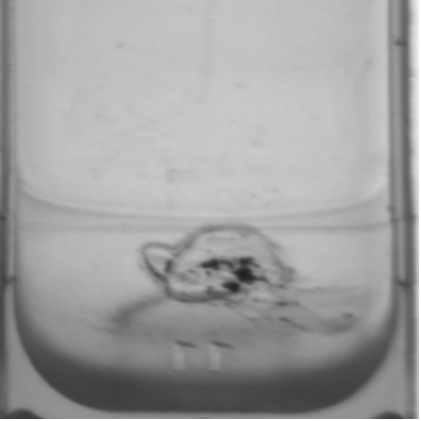}}
		\hspace{0.25cm}\subfloat[The reconstructed image $\hat{X}$]{\includegraphics[scale=0.4]{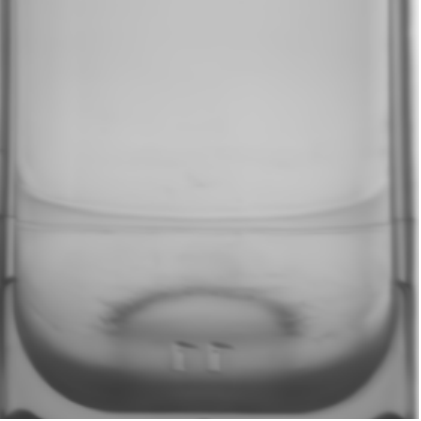}} \\
		\subfloat[The difference image]{\includegraphics[scale=0.4]{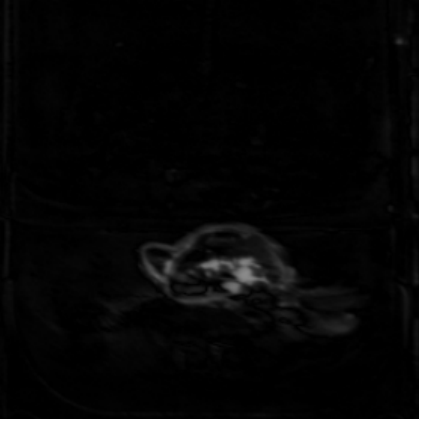}}
		\hspace{0.25cm}\subfloat[The heatmap $H$ superimposed]{\includegraphics[scale=0.4]{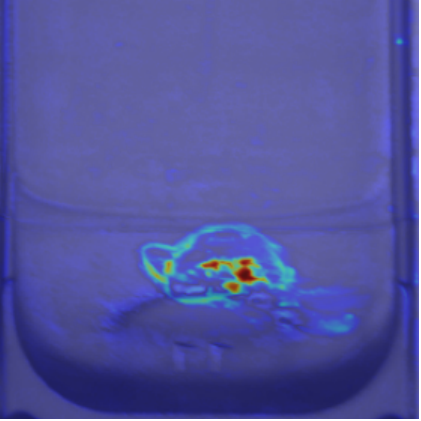}}
		\caption{Burn in the lower part of the vial.}
		\label{fig:8_ex}
	\end{figure}

	\begin{figure}[H]
		\centering
		\subfloat[Original vial region image ($X$)]{\includegraphics[scale=0.4]{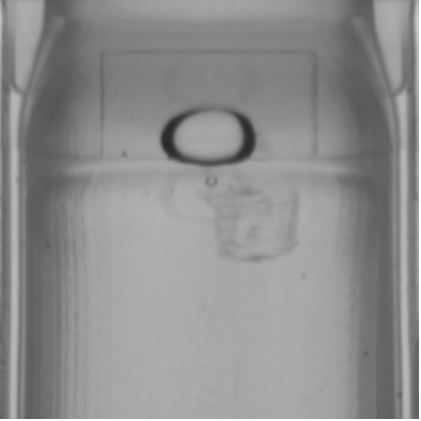}}
		\hspace{0.25cm}\subfloat[The reconstructed image $\hat{X}$]{\includegraphics[scale=0.4]{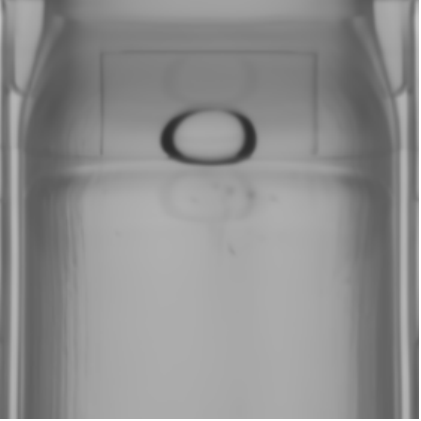}} \\
		\subfloat[The difference image]{\includegraphics[scale=0.4]{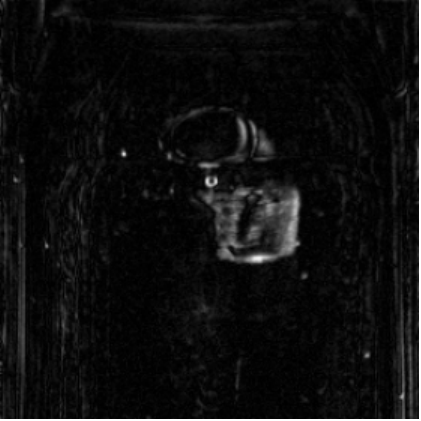}}
		\hspace{0.25cm}\subfloat[The heatmap $H$ superimposed]{\includegraphics[scale=0.4]{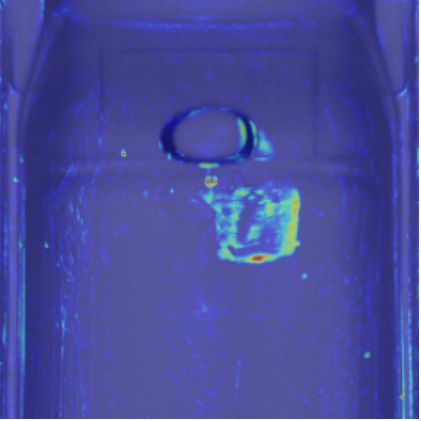}}
		\caption{Light scratch near the product neck.}
		\label{fig:9_ex}
	\end{figure}

	\begin{figure}[H]
		\centering
		\subfloat[Original vial region image ($X$)]{\includegraphics[scale=0.4]{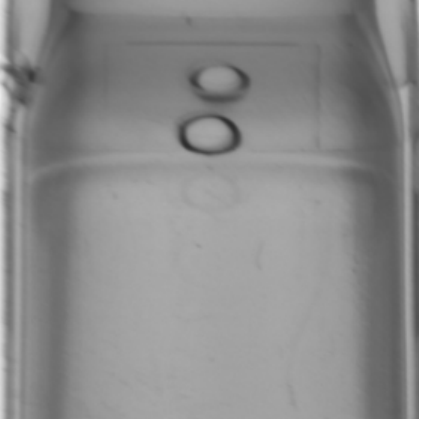}}
		\hspace{0.25cm}\subfloat[The reconstructed image $\hat{X}$]{\includegraphics[scale=0.4]{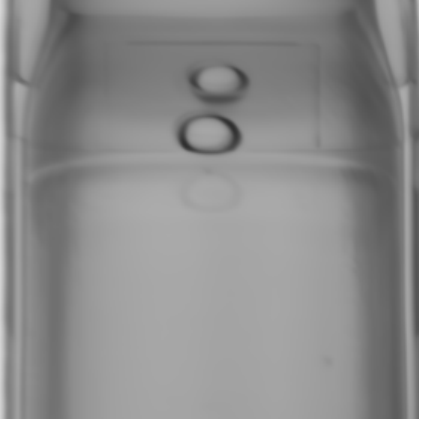}} \\
		\subfloat[The difference image]{\includegraphics[scale=0.4]{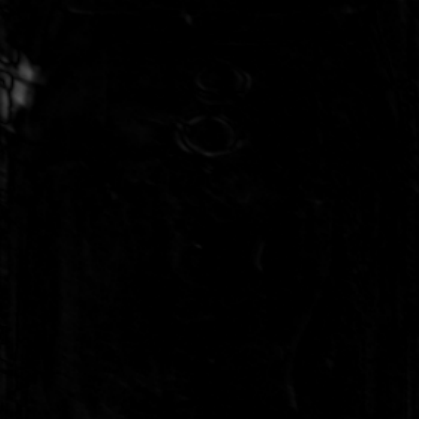}}
		\hspace{0.25cm}\subfloat[The heatmap $H$ superimposed]{\includegraphics[scale=0.4]{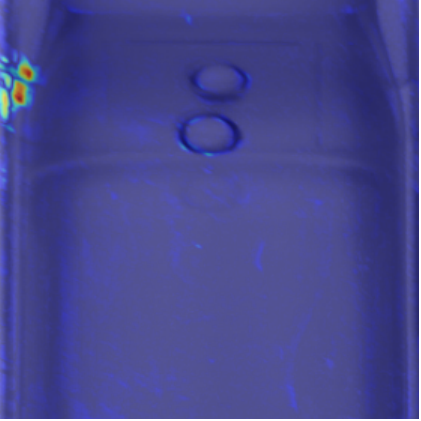}}
		\caption{External imperfection caused by laser cutting of the product.}
		\label{fig:10_ex}
	\end{figure}

	\begin{figure}[H]
		\centering
		\subfloat[Original vial region image ($X$)]{\includegraphics[scale=0.4]{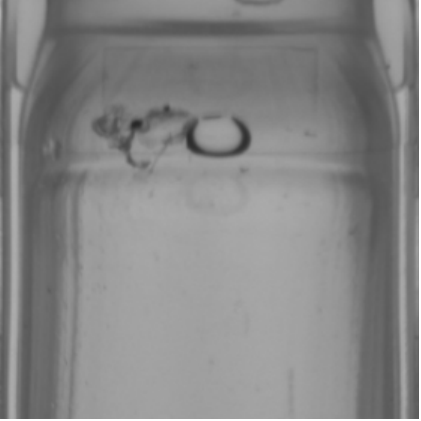}}
		\hspace{0.25cm}\subfloat[The reconstructed image $\hat{X}$]{\includegraphics[scale=0.4]{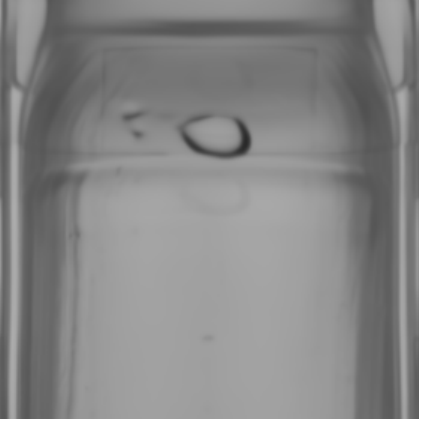}} \\
		\subfloat[The difference image]{\includegraphics[scale=0.4]{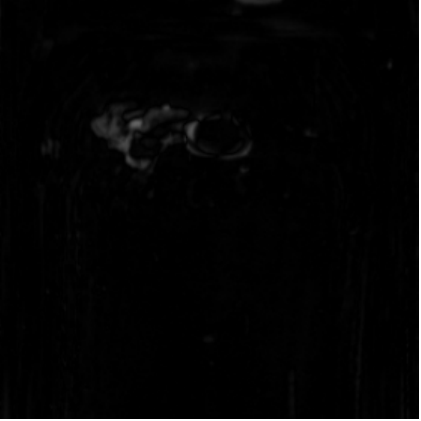}}
		\hspace{0.25cm}\subfloat[The heatmap $H$ superimposed]{\includegraphics[scale=0.4]{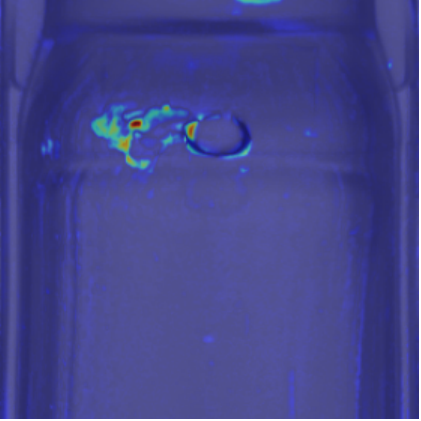}}
		\caption{Deep scratch on the upper part of the vial's body.}
		\label{fig:11_ex}
	\end{figure}

	\begin{figure}[H]
		\centering
		\subfloat[Original vial region image ($X$)]{\includegraphics[scale=0.4]{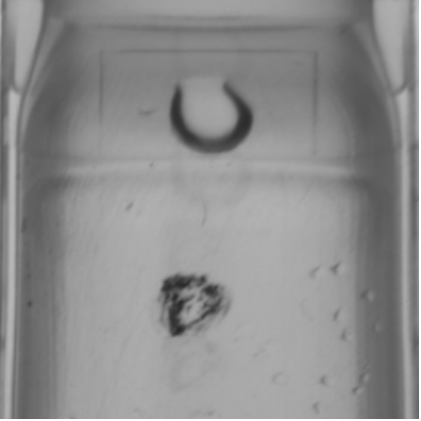}}
		\hspace{0.25cm}\subfloat[The reconstructed image $\hat{X}$]{\includegraphics[scale=0.4]{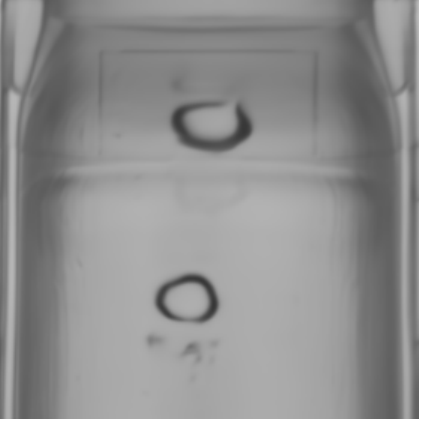}} \\
		\subfloat[The difference image]{\includegraphics[scale=0.4]{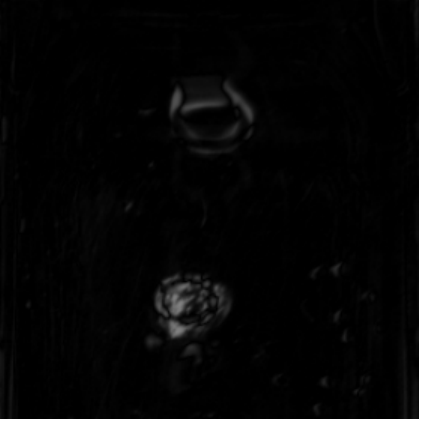}}
		\hspace{0.25cm}\subfloat[The heatmap $H$ superimposed]{\includegraphics[scale=0.4]{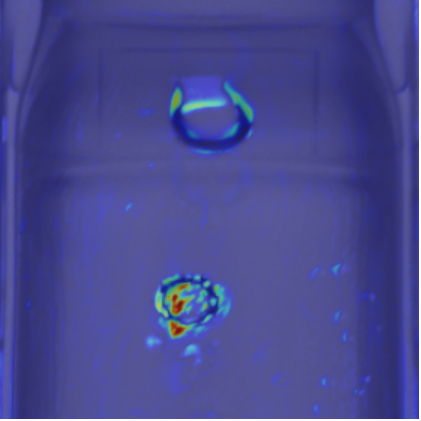}}
		\caption{Burn on the vial's body. It is reproduced as a bubble, since the network does not know how to represent the defect features.}
		\label{fig:12_ex}
	\end{figure}

	\begin{figure}[H]
		\centering
		\subfloat[Original vial region image ($X$)]{\includegraphics[scale=0.4]{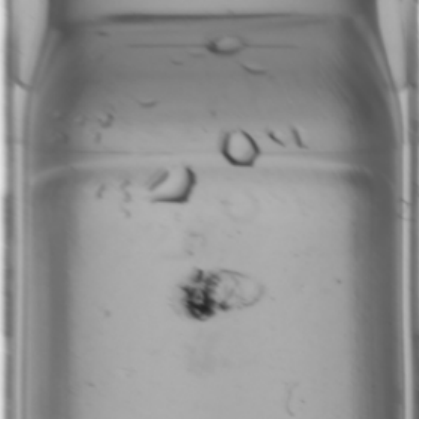}}
		\hspace{0.25cm}\subfloat[The reconstructed image $\hat{X}$]{\includegraphics[scale=0.4]{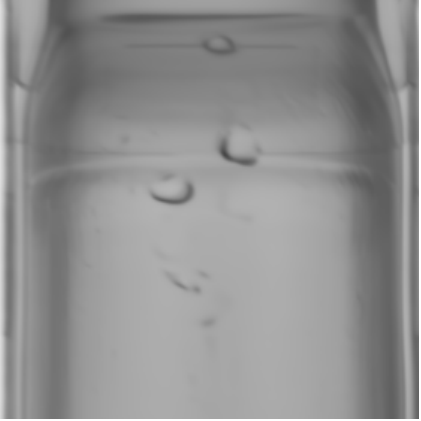}} \\
		\subfloat[The difference image]{\includegraphics[scale=0.4]{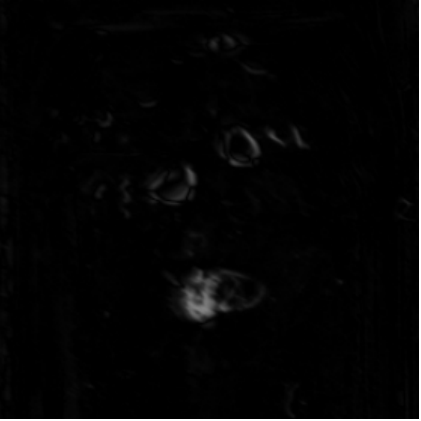}}
		\hspace{0.25cm}\subfloat[The heatmap $H$ superimposed]{\includegraphics[scale=0.4]{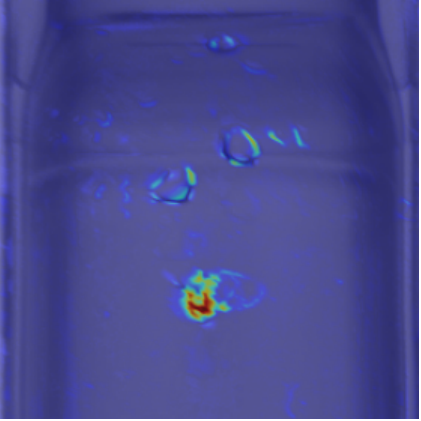}}
		\caption{Black spot on the vial's body}
		\label{fig:13_ex}
	\end{figure}

	\begin{figure}[H]
		\centering
		\subfloat[Original vial region image ($X$)]{\includegraphics[scale=0.4]{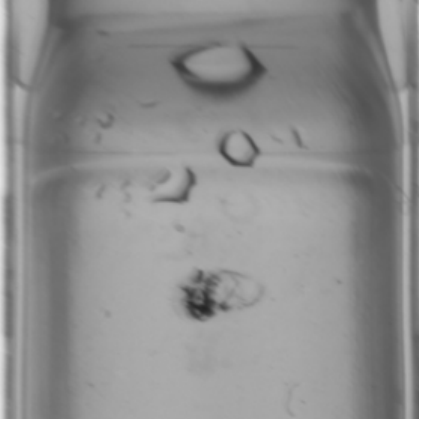}}
		\hspace{0.25cm}\subfloat[The reconstructed image $\hat{X}$]{\includegraphics[scale=0.4]{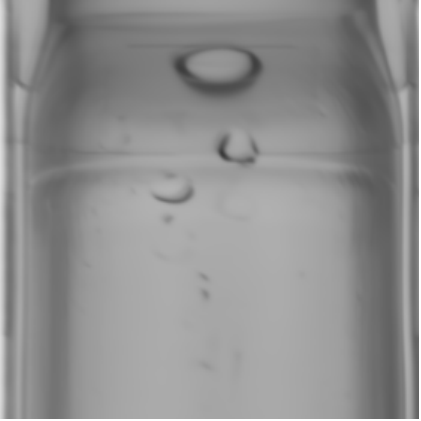}} \\
		\subfloat[The difference image]{\includegraphics[scale=0.4]{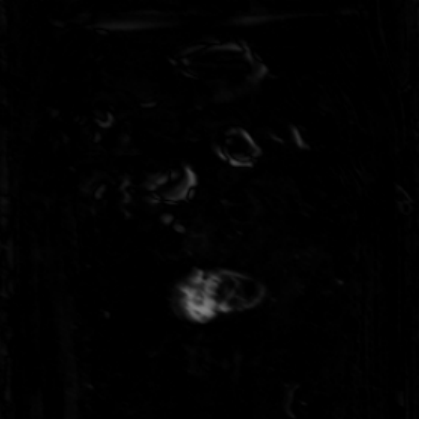}}
		\hspace{0.25cm}\subfloat[The heatmap $H$ superimposed]{\includegraphics[scale=0.4]{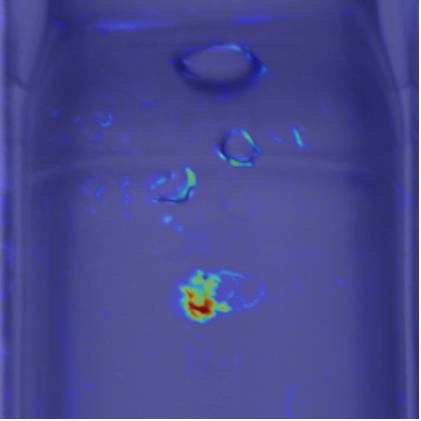}}
		\caption{Black spot of the vial's body}
		\label{fig:14_ex}
	\end{figure}

	\begin{figure}[H]
		\centering
		\subfloat[Original vial region image ($X$)]{\includegraphics[scale=0.4]{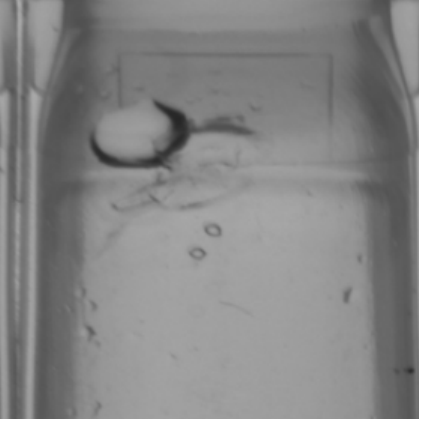}}
		\hspace{0.25cm}\subfloat[The reconstructed image $\hat{X}$]{\includegraphics[scale=0.4]{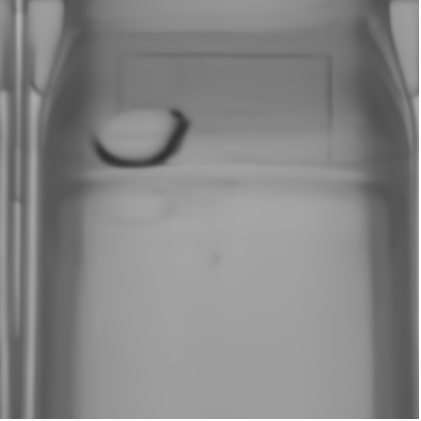}} \\
		\subfloat[The difference image]{\includegraphics[scale=0.4]{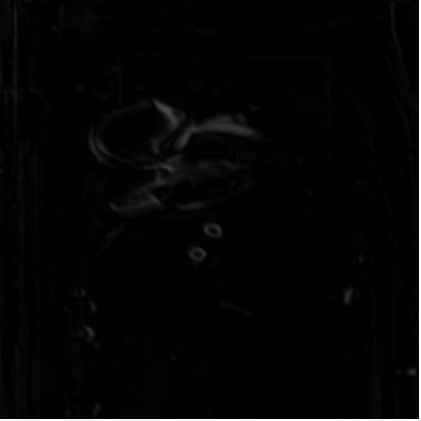}}
		\hspace{0.25cm}\subfloat[The heatmap $H$ superimposed]{\includegraphics[scale=0.4]{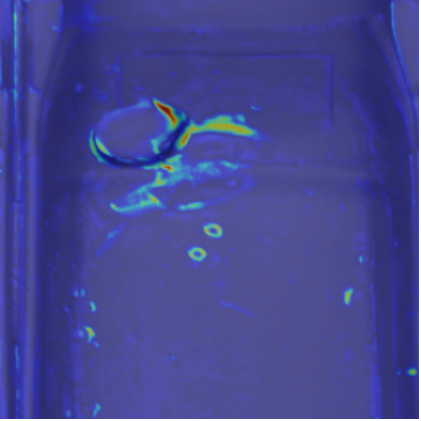}}
		\caption{Bump on the lower part of the vial's neck, near to a big bubble stuck in the internal layer of the BFS surface.}
		\label{fig:15_ex}
	\end{figure}

	\begin{figure}[H]
		\centering
		\subfloat[Original vial region image ($X$)]{\includegraphics[scale=0.4]{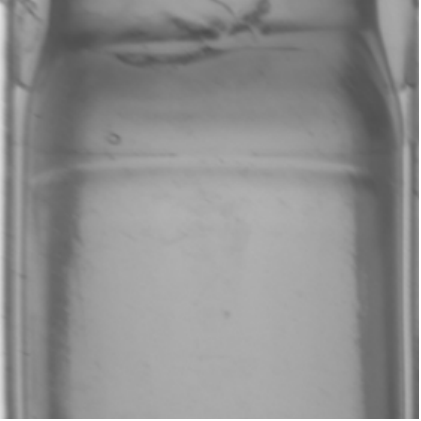}}
		\hspace{0.25cm}\subfloat[The reconstructed image $\hat{X}$]{\includegraphics[scale=0.4]{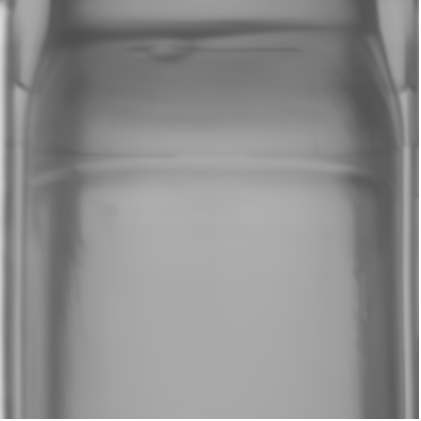}} \\
		\subfloat[The difference image]{\includegraphics[scale=0.4]{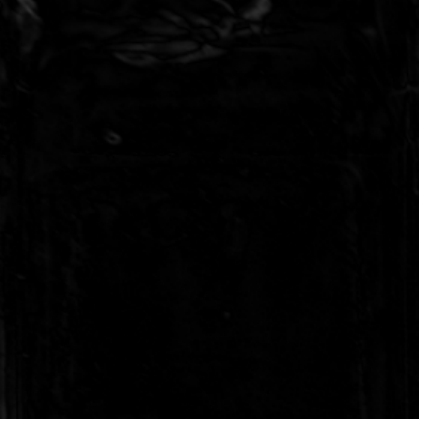}}
		\hspace{0.25cm}\subfloat[The heatmap $H$ superimposed]{\includegraphics[scale=0.4]{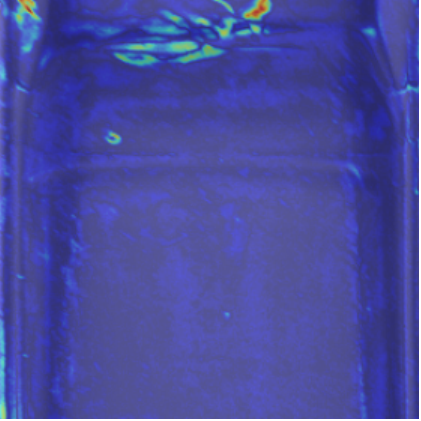}}
		\caption{Light bump on the vial's neck}
		\label{fig:16_ex}
	\end{figure}

	\begin{figure}[H]
		\centering
		\subfloat[Original vial region image ($X$)]{\includegraphics[scale=0.4]{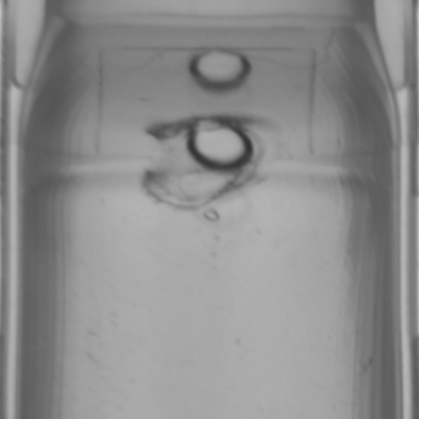}}
		\hspace{0.25cm}\subfloat[The reconstructed image $\hat{X}$]{\includegraphics[scale=0.4]{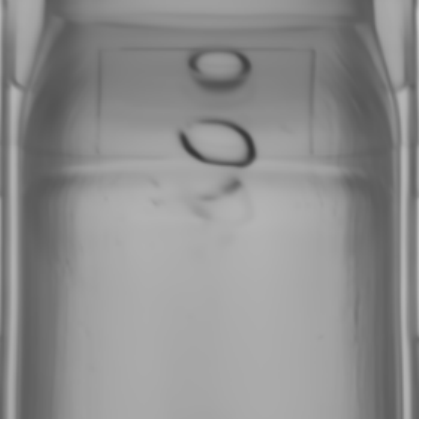}} \\
		\subfloat[The difference image]{\includegraphics[scale=0.4]{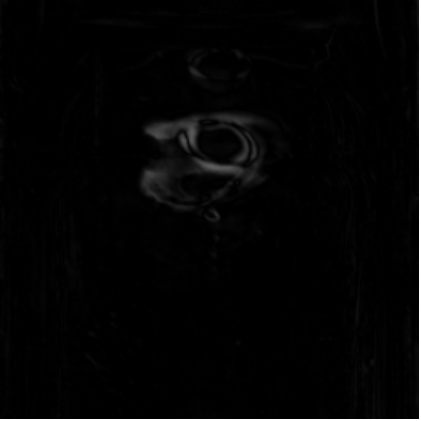}}
		\hspace{0.25cm}\subfloat[The heatmap $H$ superimposed]{\includegraphics[scale=0.4]{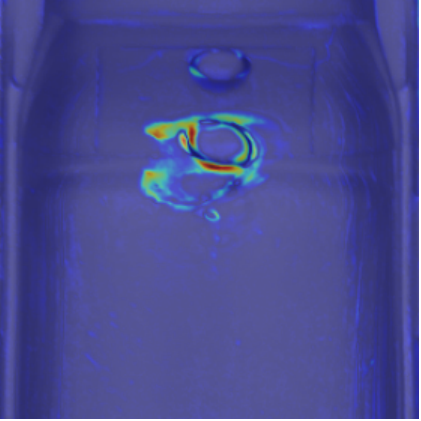}}
		\caption{Heavy deformation near the neck region, overlapped to a bubble that the network is still able to reproduce.}
		\label{fig:17_ex}
	\end{figure}

	\begin{figure}[H]
		\centering
		\subfloat[Original vial region image ($X$)]{\includegraphics[scale=0.4]{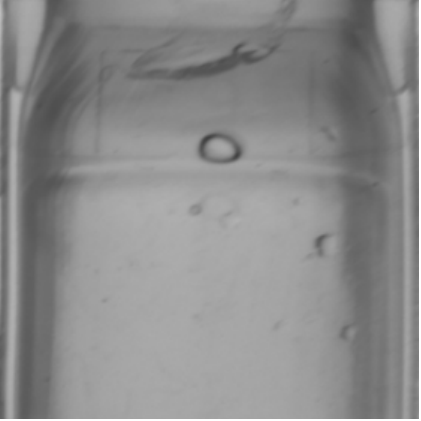}}
		\hspace{0.25cm}\subfloat[The reconstructed image $\hat{X}$]{\includegraphics[scale=0.4]{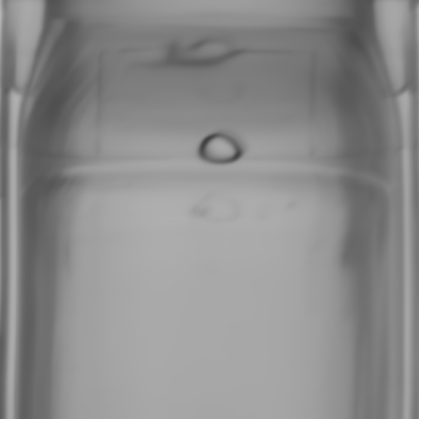}} \\
		\subfloat[The difference image]{\includegraphics[scale=0.4]{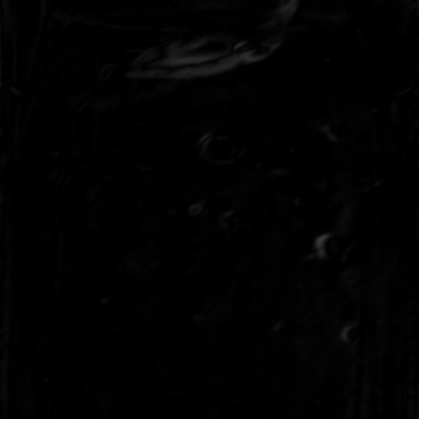}}
		\hspace{0.25cm}\subfloat[The heatmap $H$ superimposed]{\includegraphics[scale=0.4]{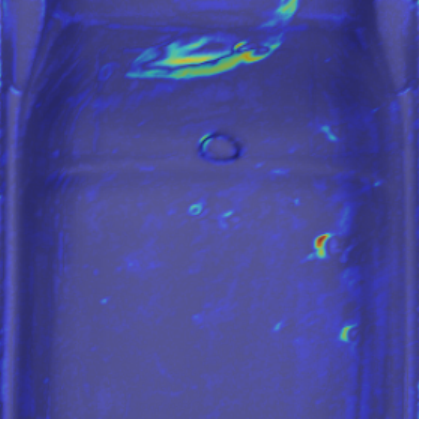}}
		\caption{Scratch on the vial's neck.}
		\label{fig:18_ex}
	\end{figure}

	\begin{figure}[H]
		\centering
		\subfloat[Original vial region image ($X$)]{\includegraphics[scale=0.4]{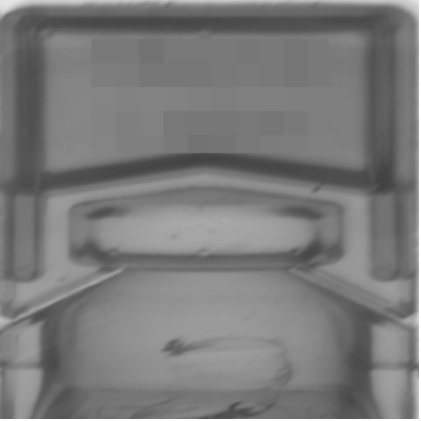}}
		\hspace{0.25cm}\subfloat[The reconstructed image $\hat{X}$]{\includegraphics[scale=0.4]{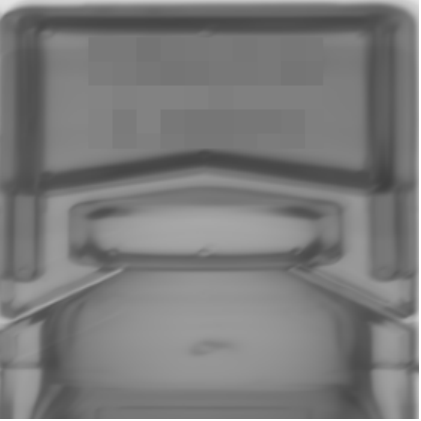}} \\
		\subfloat[The difference image]{\includegraphics[scale=0.4]{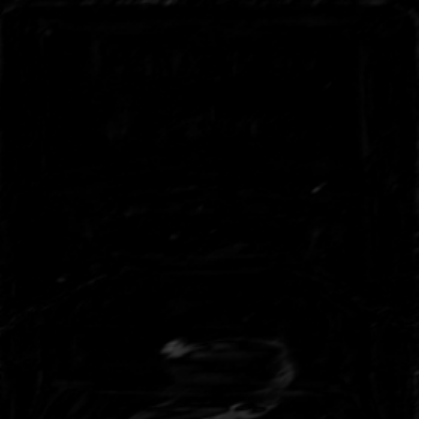}}
		\hspace{0.25cm}\subfloat[The heatmap $H$ superimposed]{\includegraphics[scale=0.4]{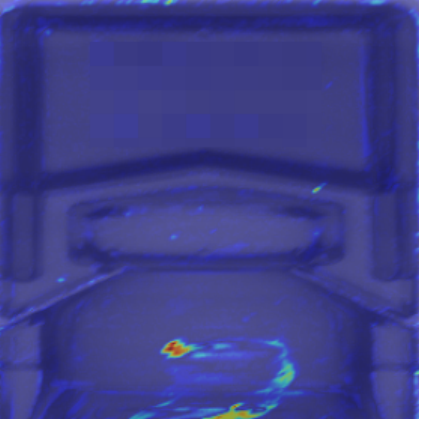}}
		\caption{Light bump on the lower part of the vial's cap region.}
		\label{fig:19_ex}
	\end{figure}

	\begin{figure}[H]
		\centering
		\subfloat[Original vial region image ($X$)]{\includegraphics[scale=0.4]{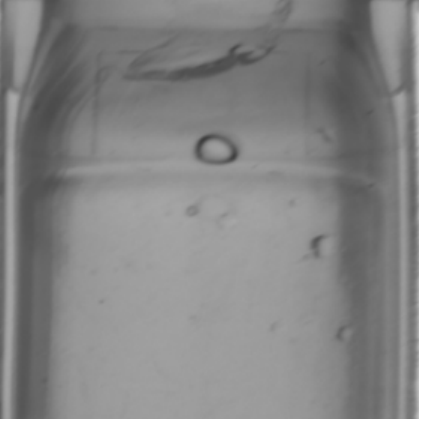}}
		\hspace{0.25cm}\subfloat[The reconstructed image $\hat{X}$]{\includegraphics[scale=0.4]{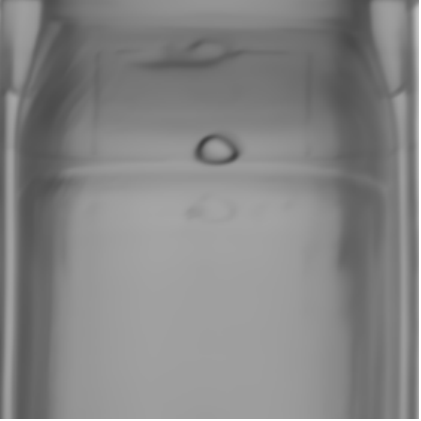}} \\
		\subfloat[The difference image]{\includegraphics[scale=0.4]{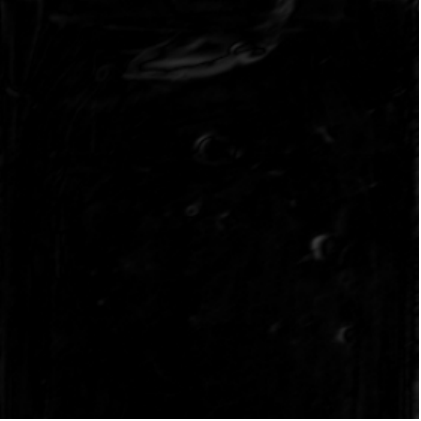}}
		\hspace{0.25cm}\subfloat[The heatmap $H$ superimposed]{\includegraphics[scale=0.4]{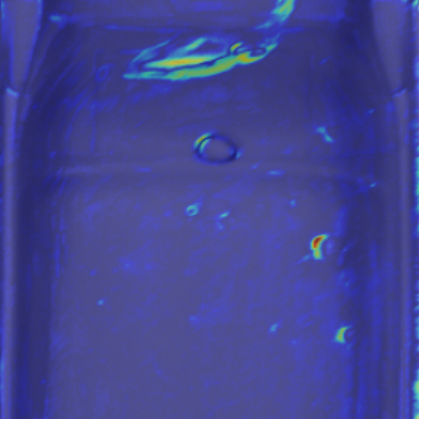}}
		\caption{Scratch on the vial's neck.}
		\label{fig:20_ex}
	\end{figure}
	}

}

\FloatBarrier
\backmatter

\bmhead{Acknowledgements}
The authors would like to thank Bonfiglioli Engineering for providing a real-case dataset to test the software developed in this work.
The first author is supported by an industrial PhD funded by Bonfiglioli Engineering, Ferrara, Italy.
Alice Bizzarri is supported by a National PhD funded by Politecnico di Torino, Torino, Italy and Università di Ferrara, Ferrara, Italy.

\section*{Declarations}
\noindent\textbf{Funding}
The first author is supported by an industrial PhD funded by Bonfiglioli Engineering, Ferrara, Italy. Alice Bizzarri is supported by a National PhD funded by Politecnico di Torino, Torino, Italy and Università di Ferrara, Ferrara, Italy.

\medskip
\noindent\textbf{Conflict of interest/Competing interests}
The authors declare that they have no competing interests.

\medskip
\noindent\textbf{Ethics approval and consent to participate}
Not applicable.

\medskip
\noindent\textbf{Consent for publication}
Not applicable.

\medskip
\noindent\textbf{Data availability}
The datasets generated and/or analyzed during the current study are not publicly available due to NDA and commercial confidentiality constraints, but may be available from the corresponding author on reasonable request and where contractually permitted.

\medskip
\noindent\textbf{Materials availability}
Not publicly available due to NDA and commercial confidentiality constraints.

\medskip
\noindent\textbf{Code availability}
Not publicly available due to NDA and proprietary constraints.

\medskip
\noindent\textbf{Author contribution}
All authors contributed to the study conception and design. Material preparation, data collection, analysis, and manuscript drafting/revision were performed by the authors. All authors read and approved the final manuscript.

\bibliography{biblio}

\end{document}